\documentclass[journal]{IEEEtran}
\usepackage[T1]{fontenc}
\usepackage{graphicx}
\usepackage{amsmath}
\usepackage{amssymb}
\usepackage{amsfonts}
\usepackage{mathtools}
\usepackage{nicefrac}
\usepackage{booktabs}
\usepackage{multirow}
\usepackage{makecell}
\usepackage{array}
\usepackage{subcaption}
\usepackage{xcolor}
\usepackage{comment}
\usepackage{tikz}
\usetikzlibrary{arrows.meta, positioning, calc}
\usepackage{stfloats}   % allows double-column floats at the bottom of a page
\usepackage{url}
\usepackage{cite}

\usepackage{amsthm}

\usepackage[hidelinks]{hyperref}
\usepackage[capitalize,noabbrev]{cleveref}

\theoremstyle{plain}
\newtheorem{theorem}{Theorem}
\newtheorem{proposition}[theorem]{Proposition}

\newtheorem{corollary}[theorem]{Corollary}
\theoremstyle{definition}

\theoremstyle{remark}
\newtheorem{remark}[theorem]{Remark}

\begin{document}

\title{Neural Bridge Processes}

\author{Jian~Xu,
        Yican~Liu,
        Delu~Zeng,
        John~Paisley,
        and~Qibin~Zhao%
\thanks{J. Xu and Q. Zhao are with the RIKEN Center for Advanced Intelligence
Project (AIP), Tokyo, Japan (e-mail: jian.xu@riken.jp; qibin.zhao@riken.jp).}%
\thanks{Y. Liu and D. Zeng are with the South China University of Technology,
Guangzhou, China (e-mail: liuyc57@chinatelecom.cn; dlzeng@scut.edu.cn).}%
\thanks{J. Paisley is with Columbia University, New York, NY, USA
(e-mail: jwp2128@columbia.edu).}%
\thanks{Corresponding author: Jian Xu.}}

% Preprint: no journal running head.
\markboth{}{}

\maketitle

\begin{abstract}
Learning stochastic functions from partially observed context-target pairs requires models that are expressive, uncertainty-aware, and strongly conditioned on inputs. Neural Diffusion Processes (NDPs) improve expressivity with denoising diffusion, but their forward process is input-independent; inputs only enter the reverse denoiser, so the noisy training states themselves do not encode the conditioning inputs. We propose Neural Bridge Processes (NBPs), which replace the unconditional forward kernel with an input-anchored bridge trajectory. When input and output dimensions differ, NBP learns an output-space anchor \(a_\psi(x)=P_\psi(x)\), allowing coordinates or other inputs to guide the generative path without changing the denoising backbone. We show that process-level anchoring (i) admits a path-measure KL identifiability bound with rate $\sum_t\gamma_t^2/\beta_t$, (ii) makes the forward state Fisher-informative about \(x\) and yields a Cram\'er--Rao limit on input recovery from the trajectory, and (iii) opens a second, input-mediated pathway through which the loss responds to the input, whose magnitude is governed by the same Fisher rate. These identities characterize what the anchored trajectory retains about the input; they concern sensitivity to the input rather than the parameter gradients that training actually follows, and they do not by themselves imply lower predictive error, which we establish empirically. Experiments on Gaussian-process regression up to six input dimensions, Bayesian optimization, EEG, CylinderFlow, and image regression show consistent improvements over the corresponding unanchored diffusion process, without loss of sample diversity or calibration. Ablations attribute the gains to the full bridge construction with learned alignment, and the same input-anchored path principle transfers to Flow Matching Neural Processes. Bridge-anchored generative paths are thus a useful design principle for conditional stochastic function modeling in the Neural Process setting.

\end{abstract}

\section{Introduction}

\IEEEPARstart{L}{earning} stochastic functions from partially observed context-target pairs is a fundamental problem in probabilistic modeling~\cite{rasmussen2003gaussian,garnelo2018neural, garnelo2018conditional,dutordoir2023neural,franzese2023continuous, bonito2023diffusion, mathieu2023geometric,chang2024amortized,dou2025score,hamadflow}, with applications in meta-learning~\cite{garnelo2018neural, garnelo2018conditional}, few-shot regression~\cite{kim2019attentive}, Bayesian optimization~\cite{dutordoir2023neural,krishnamoorthy2023diffusion}, and uncertainty-aware prediction~\cite{chang2024amortized,dou2025score,hamadflow, requeima2024llm}. These settings require models that generalize across tasks while remaining calibrated under scarce or incomplete observations.
Gaussian Processes (GPs)~\cite{rasmussen2003gaussian} have traditionally dominated this area due to their analytical tractability and clear uncertainty quantification. However, GPs inherently assume Gaussianity and exhibit cubic computational complexity with respect to data size, severely limiting their applicability in scenarios involving large datasets or inherently non-Gaussian functional distributions~\cite{snelson2005sparse,titsias2009variational,xu2024sparse,xu2024sparse1}.

Neural Processes (NPs)~\cite{garnelo2018neural, garnelo2018conditional,kim2019attentive, louizos2019functional} offer a scalable alternative by amortizing stochastic function prediction with neural architectures. However, standard NPs often struggle to capture complex multi-modal target distributions, motivating stronger generative mechanisms.

NDPs~\cite{dutordoir2023neural} address this by modeling the input-output mapping as a learned diffusion process~\cite{ho2020denoising}, offering enhanced expressivity and sample diversity. Despite their promise in stochastic function modeling, NDPs rely on an unconditional forward process. The input affects the reverse denoiser, but the noised state \(y_t\) used during training is sampled from a distribution that does not depend on \(x\). This creates a structural gap for conditional function modeling: input information is supplied only through the learned reverse network, rather than through the generative path itself.

In this work, we introduce Neural Bridge Processes (NBP), a diffusion-based generative framework that integrates input supervision throughout the diffusion trajectory. NBPs reformulate the forward diffusion kernel so that intermediate noisy states are dynamically anchored by the input through a bridge coefficient $\gamma_t$, together with a reverse correction term that keeps the reverse dynamics consistent with the modified forward process. When input and output dimensions differ, such as in coordinate-to-RGB image regression, we use a learned output-space anchor $a_\psi(x)=P_\psi(x)$ while leaving the denoising architecture unchanged; details and a visualization are provided in Sec.~\ref{3.5}.

NBP is related to conditional diffusion and diffusion bridge models, but differs in both the object being modeled and the location of conditioning. It learns distributions over functions and places the input anchor directly in the DDPM-style forward kernel, rather than only conditioning the denoiser or constructing a generic data-to-data bridge~\cite{zhou2023denoising, yue2023image, zheng2024diffusion, li2023bbdm,peluchetti2023diffusion,he2024consistency,shi2023diffusion,naderiparizi2025constrained}.

We provide a theoretical analysis showing that process-level anchoring gives NBP properties unavailable to standard NDPs: the bridge construction admits a path-measure KL identifiability bound at rate $\sum_t\gamma_t^2/\beta_t$, the forward state carries an explicit Fisher information matrix about $x$, yielding a Cram\'er--Rao limit on any unbiased estimator of the input from the trajectory, and the derivative of the denoising loss with respect to the input acquires a second, input-mediated term whose magnitude is governed by the same per-step Fisher rate $\bar{\gamma}_t^2/(1-\bar{\alpha}_t)$. None of these properties hold for NDPs, where the corresponding quantities are identically zero. We emphasize what these results do \emph{not} say: they establish that the NBP trajectory retains and exposes information about $x$, but retaining information is not the same as using it well, and none of the identities implies a lower predictive error on a given task. The intended benefit is a finite-capacity inductive bias, which is an empirical claim, and we treat it as such throughout. We validate NBPs on Gaussian-process regression up to six input dimensions, Bayesian optimization, electroencephalogram (EEG) regression~\cite{zhang1995event}, CylinderFlow, and image regression. Sec.~\ref{sec:celeba_component_ablation} isolates the source of the gains through component ablations, Sec.~\ref{sec:bimodal} identifies a regime in which the anchor supervision is inappropriate, and Sec.~\ref{sec:flow-bridge-ablation} shows that the same bridge idea transfers to Flow Matching Neural Processes (FlowNP)~\cite{hamadflow}.

In summary, the core contributions of this paper are: 
\begin{itemize}
\item We introduce \textbf{Neural Bridge Processes (NBPs)}, a class of stochastic function models that use input-anchored diffusion trajectories to make the noisy training states explicitly depend on the conditioning inputs.
\item We formulate a DDPM-style bridge for function modeling, including a learned output-space anchor $a_\psi(x)=P_\psi(x)$ for dimension-mismatched inputs and outputs.
\item We provide a theoretical analysis of process-level anchoring with three quantitative results: a path-measure KL identifiability bound with rate $\sum_t\gamma_t^2/\beta_t$; a Fisher information identity for the forward bridge that yields a Cram\'er--Rao lower bound on input recoverability from the trajectory; and an input-sensitivity analysis tying the SNR-aware schedule to the per-step Fisher rate.
\item We validate NBPs on Gaussian-process regression up to six input dimensions, Bayesian optimization, EEG, CylinderFlow, and image regression, under a single controlled protocol in which NDP and NBP differ only through the bridge flag. Ablations---including anchor-fairness controls that give NDP the same auxiliary $P_\psi$ head and loss but no forward bridge---attribute the gains to the bridge construction rather than to architecture size, denoiser capacity, or the extra coordinate-to-output supervision, while a test-time intervention shows that a trained model's predictions depend on the deployed bridge scale, and diversity/calibration diagnostics show that the gains do not come from reduced sample dispersion.
\item We delimit the construction: we identify a multimodal regime in which the deterministic anchor supervision is inappropriate, and we state the assumptions under which each theoretical identity holds, including the generalization to arbitrary additive Gaussian bridges and the cases it excludes. We further show that the same bridge idea can improve a Flow Matching Neural Process.
\end{itemize}

\section{Related Work}

Neural Processes (NPs)~\cite{garnelo2018neural,garnelo2018conditional,kim2019attentive,louizos2019functional} learn distributions over functions from context-target observations, combining amortized neural prediction with uncertainty modeling. A broad line of work has improved NP expressivity through attention, convolutional structure, autoregressive modeling, graph structure, and other process-aware architectures~\cite{gordon2019convolutional,foong2020meta,singh2019sequential,nguyen2022transformer,bruinsma2023autoregressive,hu2023graph}. More recent methods incorporate stronger generative mechanisms into function modeling, including diffusion-based neural processes~\cite{dutordoir2023neural,mathieu2023geometric}, score-based neural processes~\cite{dou2025score}, flow-based models~\cite{hamadflow}, and related neural ODE formulations~\cite{chen2018neural,norcliffe2021neural}.

Our work is closest to Neural Diffusion Processes (NDPs)~\cite{dutordoir2023neural}, which are already a conditional diffusion baseline for Neural Process-style function learning: they learn a DDPM-style reverse denoiser conditioned on the input and context observations. The key difference is that NDPs retain an unconditional forward diffusion process, so the noised training states are generated independently of the input. NBP keeps the same conditional denoising backbone but modifies the forward transition itself, so the input acts as a dynamic anchor throughout the diffusion trajectory. This process-level modification distinguishes NBP from standard conditional diffusion models~\cite{choi2021ilvr,zhang2023adding,zhu2023conditional}, where conditioning typically enters the reverse model while the forward noising process remains condition-independent.

NBP is also related to diffusion bridge models~\cite{zhou2023denoising,yue2023image,zheng2024diffusion,li2023bbdm,peluchetti2023diffusion,he2024consistency,shi2023diffusion,naderiparizi2025constrained}. Diffusion bridges have become popular in generative modeling, but existing DDBM-style methods are usually designed for data-space transport between paired endpoint distributions, such as image-to-image translation. This differs from the Neural Process setting, where the goal is to model stochastic functions from context-target observations and predict outputs at arbitrary target inputs. We do not claim the bridge principle itself as a new mathematical concept: diffusion bridges, Schr\"odinger bridges and data-to-data transport are established tools. Our contribution is its specialization to stochastic function modeling, where the conditioning variable is a function input rather than a paired data-space endpoint: an input-anchored DDPM forward kernel for conditional function prediction, a learned output-space anchor for $D_x\neq D_y$, the corresponding reverse correction derived from the modified forward process, and the characterization of this construction through path-measure KL, forward-state Fisher information and the additional gradient pathway. Sec.~\ref{sec:flow-bridge-ablation} further compares this bridge principle with recent flow-based Neural Processes.

\section{Preliminaries}
Neural Processes (NPs) approximate a stochastic process \( F: X \to Y \) through finite-dimensional marginals. Given context observations \( (x_\mathcal{C}, y_\mathcal{C}) \) and target inputs \( x_\mathcal{T} \), a latent-variable NP predicts target outputs by
\begin{equation}
p(y_\mathcal{T}, z|x_\mathcal{T}, x_\mathcal{C}, y_\mathcal{C}) = p(z|x_\mathcal{C}, y_\mathcal{C}) \prod_{i=1}^{|\mathcal{T}|} p(y_{\mathcal{T},i}|x_{\mathcal{T},i}, z)
\end{equation}
where \(z\) captures uncertainty about the global structure of the underlying function. NDPs replace simple latent-variable decoders with a learned diffusion process over outputs, which improves expressivity but still uses an unconditional forward noising process.

\subsection{Neural Diffusion Processes}

Given clean outputs $y_0$, NDPs~\cite{dutordoir2023neural} define the standard DDPM forward process
\begin{equation}
\begin{aligned}
q(y_{1:T} \mid y_0) =& \prod_{t=1}^T q(y_t \mid y_{t-1}),
\\q(y_t \mid y_{t-1}) =& \mathcal{N}(y_t; \sqrt{1 - \beta_t} \, y_{t-1}, \beta_t I).\end{aligned}
\end{equation}
The terminal variable is therefore approximately standard Gaussian noise and does not encode the conditioning input. The reverse process is conditioned on $x$ through a denoising network:
\begin{equation}  
p_\theta(y_{0:T} \mid x) = p(y_T) \prod_{t=1}^T p_\theta(y_{t-1} \mid y_t, x),  
\end{equation}  
where $p_\theta(y_{t-1} \mid y_t, x)$ is parameterized by a noise predictor $\epsilon_\theta(y_t,t,x)$. Training uses the usual denoising objective
\begin{equation}\begin{aligned}
\mathcal{L}_\theta =& \mathbb{E}_{t, x, y_0, \epsilon} \left[ \left\| \epsilon - \epsilon_\theta(y_t, t, x) \right\|_2^2 \right],
\\ \text{with} \quad y_t = &\sqrt{\bar{\alpha}_t} y_0 + \sqrt{1 - \bar{\alpha}_t} \epsilon.\end{aligned}
\end{equation}
Here and throughout, $\alpha_t=1-\beta_t$ and $\bar{\alpha}_t=\prod_{s=1}^{t}\alpha_s$ denote the usual one-step and cumulative signal-retention coefficients of the variance schedule, so that $\bar{\alpha}_0=1$ and $\bar{\alpha}_t$ decreases monotonically towards $0$. This compact review highlights the key design choice that motivates NBP: in NDPs, $x$ affects the reverse denoiser but not the forward trajectory.

\section{Method: Neural  Bridge Processes (NBP)}

\subsection{Problem Setup}
\label{4.1}
We consider the standard meta-learning setting where a model observes a set of context points $x_\mathcal{C} = \{(x_i, y_i)\}_{i=1}^{N_c}$ and aims to predict the corresponding outputs $y_\mathcal{T}=\{y_j\}_{j=1}^{N_t}$ for a set of target inputs $x_\mathcal{T}=\{x_j\}_{j=1}^{N_t}$. Here, each task is assumed to be sampled from a distribution over functions, and the goal is to model the conditional distribution $p(y_\mathcal{T} | x_\mathcal{C}, x_\mathcal{T},y_\mathcal{C})$.  
Throughout the method section, we use $x$ and $y$ to denote the full set of inputs and outputs for one task-level function, while $(x_\mathcal{C},y_\mathcal{C})$ and $(x_\mathcal{T},y_\mathcal{T})$ denote the context and target subsets used for conditional prediction. We use calligraphic symbols \(\mathcal{C}\) and \(\mathcal{T}\) for context and target index sets to avoid confusion with the complex numbers. The diffusion process is defined on the full task-level output set $y_{0:T}$ conditioned on the corresponding inputs $x$; the context--target partition is then instantiated during conditional sampling.
%Following the Neural Processes (NP) framework ~\cite{garnelo2018neural, garnelo2018conditional,kim2019attentive}, we seek a stochastic model that can flexibly predict outputs given arbitrary context-target splits, while providing calibrated uncertainty estimates.

\subsection{Motivation}
\label{3.3}

In NDPs, the input $x$ enters only through the learned denoiser $\epsilon_\theta(y_t,x,t)$; the forward kernel $q(y_t\mid y_0)$ is unconditional, so noisy training states are generated independently of $x$. NBP closes this gap by injecting $x$ into the forward transition itself, so the coupling between $x$ and $y_t$ is present throughout the trajectory rather than only in the reverse model.

\emph{Why should this help?} Input-independence of the forward process is not a defect in the infinite-capacity limit: a sufficiently expressive denoiser can in principle recover the conditional reverse dynamics from its $x$-conditioning branch alone. The argument for anchoring is about finite capacity. In NDP the corruption kernel is identical for every input, so as the noise level grows the noisy state retains progressively less input--output structure and the denoiser must reconstruct the conditional dynamics entirely through its separate conditioning path, precisely where the clean-output signal is weakest. NBP instead centres the noisy state on an input-dependent anchor,
\begin{equation}
y_t=\sqrt{\bar{\alpha}_t}\,y_0+\bar{\gamma}_t\,a_\psi(x)+\sqrt{1-\bar{\alpha}_t}\,\epsilon,
\end{equation}
so a finite-capacity denoiser learns residual refinement around an $x$-aligned trajectory, and the SNR-aware schedule makes this guidance strongest exactly when the clean signal is weakest. This is an inductive-bias argument, not a performance theorem: it predicts a benefit when the anchor carries task-relevant structure, and no benefit---or harm---when it does not. The controlled evidence is correspondingly empirical. Giving NDP the same anchor head and auxiliary loss without modifying the forward path yields only a small improvement, whereas the complete bridge produces substantially larger gains (Sec.~\ref{sec:celeba_component_ablation}); and a test-time intervention that rescales only the deployed bridge coefficient of a single trained model traces a single-peaked response around the trained scale (Table~\ref{tab:gamma_intervention}), indicating that the trained model's predictions do depend on the bridge path.

As a diagnostic of effective input coupling---in the spirit of Cram\'er--Rao sensitivity limits for parameter resolvability~\cite{zmuidzinas2003cramer}---we measure the Jacobian sensitivity
\begin{equation}
S(t)=\mathbb{E}\left[\left\|\frac{\partial \epsilon_\theta(y_t,x,t)}{\partial x}\right\|_F\right],
\end{equation}
averaged over 100 CelebA $32\times32$ test images at context ratio 0.3 (details in Supplementary Sec.~\ref{sec:sensitivity}). Figure~\ref{fig:sensitivity_curve} shows that NBP has substantially higher early-timestep sensitivity than NDP, because $x$ also enters the forward state via the bridge term. This is a \emph{descriptive} diagnostic of how the two learned functions differ before de-shifting: it shows that the trained NBP denoiser uses the input rather than ignoring it, but it does not establish that greater sensitivity improves prediction, and we do not present it as evidence of benefit. The quantitative evidence comes from likelihood, MSE, calibration, and the ablations above.

\begin{figure}[t]
    \centering
    \includegraphics[width=0.95\columnwidth]{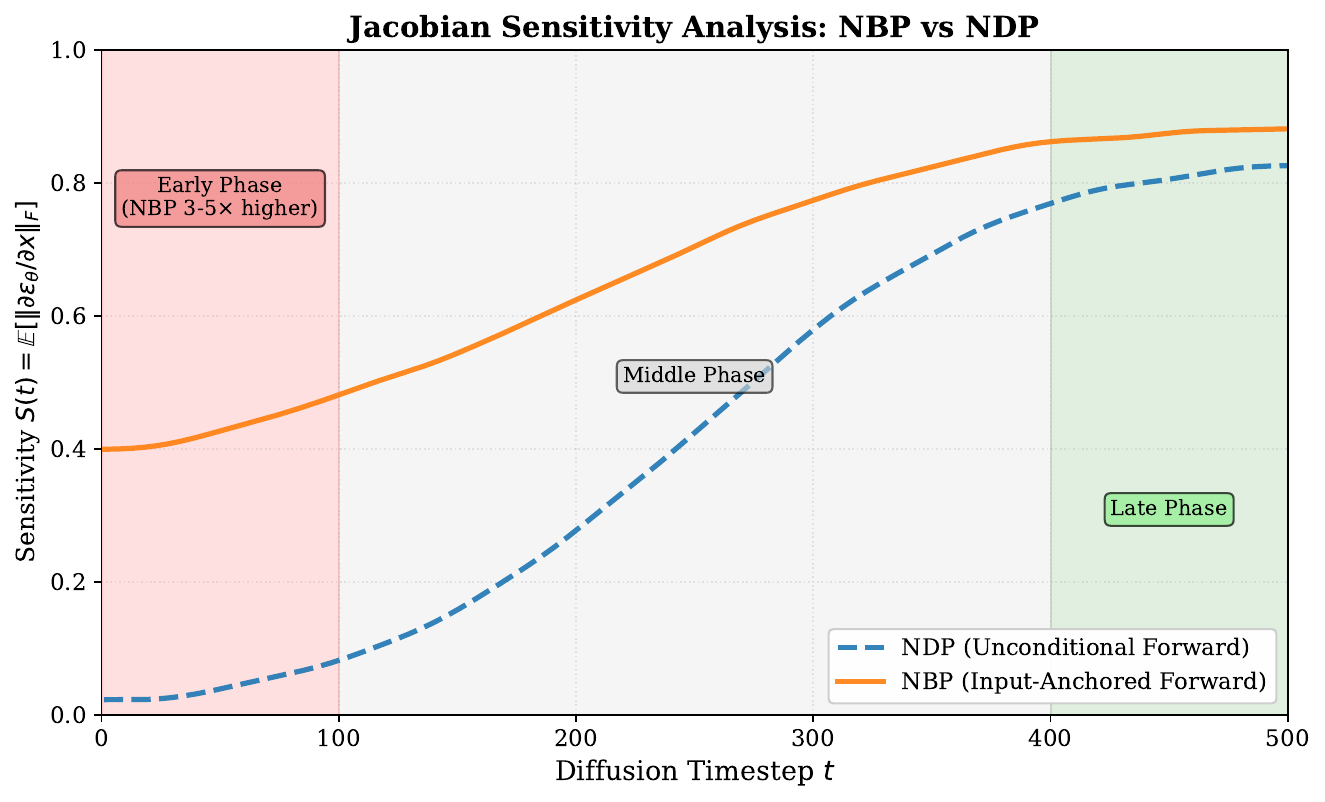}
    \caption{Jacobian sensitivity across diffusion timesteps on CelebA \(32\times32\) with context ratio 0.3, averaged over 100 test images. NBP shows larger early-timestep sensitivity than NDP.}
    \label{fig:sensitivity_curve}
\end{figure}

\subsection{Bridge Construction }

Given observed pairs $(x,y)$ with $x\in\mathbb{R}^{N\times D_x}$, $y_0=y\in\mathbb{R}^{N\times D_y}$, and the learned anchor map $a_\psi:\mathbb{R}^{D_x}\to\mathbb{R}^{D_y}$ of Sec.~\ref{3.5} (with $a_\psi=\mathrm{id}$ when $D_x=D_y$), NBP replaces the unconditional DDPM forward transition with
\begin{equation}
\label{eq7}
q(y_t|y_{t-1},x) = \mathcal{N}\!\left(y_t;\,\underbrace{\sqrt{1-\beta_t}\,y_{t-1} + \gamma_t\,a_\psi(x)}_{\text{Bridge-anchored mean}},\,\beta_t I\right).
\end{equation}
\begin{figure*}[!t]
\centering
\resizebox{0.95\textwidth}{!}{%
\begin{tikzpicture}[
    font=\small,
    node distance=0.9cm,
    box/.style={draw, rounded corners, align=center, minimum height=0.8cm, minimum width=2.4cm, fill=gray!8},
    bluebox/.style={draw, rounded corners, align=center, minimum height=0.8cm, minimum width=2.6cm, fill=blue!8},
    greenbox/.style={draw, rounded corners, align=center, minimum height=0.8cm, minimum width=2.8cm, fill=green!8},
    orangebox/.style={draw, rounded corners, align=center, minimum height=0.85cm, minimum width=3.2cm, fill=orange!12},
    arrow/.style={-Latex, thick},
    dashedarrow/.style={-Latex, thick, dashed}
]
\node[box] (x) {$x$\\input coordinate\\$\mathbb{R}^{D_x}$};
\node[bluebox, right=1.0cm of x] (pe) {positional\\encoding};
\node[greenbox, right=1.0cm of pe] (anchor) {anchor head\\$a_\psi(x)=P_\psi(x)$\\$\mathbb{R}^{D_y}$};
\node[box, below=1.0cm of anchor] (y0) {clean output\\$y_0\in\mathbb{R}^{D_y}$};
\node[box, left=1.0cm of y0] (noise) {noise\\$\epsilon$};
\node[orangebox, right=1.05cm of anchor] (bridge) {bridge state\\$y_t=\sqrt{\bar\alpha_t}y_0+\bar\gamma_t a_\psi(x)$\\$+\sqrt{1-\bar\alpha_t}\epsilon$};
\node[box, right=1.05cm of bridge] (denoiser) {denoiser\\$\epsilon_\theta(y_t,x,t)$};
\node[box, below=0.9cm of denoiser] (loss) {training losses\\denoising + anchor};

\draw[arrow] (x) -- (pe);
\draw[arrow] (pe) -- (anchor);
\draw[arrow] (anchor) -- (bridge);
\draw[arrow] (y0) -- (bridge);
\draw[arrow] (noise) -- (bridge);
\draw[arrow] (bridge) -- (denoiser);
\draw[dashedarrow] (x) to[bend left=18] node[above, align=center] {raw input remains\\denoiser conditioning} (denoiser);
\draw[arrow] (denoiser) -- (loss);
\draw[arrow] (anchor) to[bend right=18] node[right, align=center] {$\|a_\psi(x)-y_0\|^2$} (loss);
\end{tikzpicture}
}
\caption{Learned input-output alignment for NBP when $D_x\neq D_y$. The raw input remains denoiser conditioning, while $P_\psi$ maps it into the output space before it enters the bridge trajectory. In image regression, this maps 2D pixel coordinates to RGB-valued anchors.}
\label{fig:learned_anchor_alignment}
\end{figure*}

The scalar $\gamma_t$ controls the strength of the input anchor. In this sense, the anchor is dynamic: the bridge term $\gamma_t\,a_\psi(x)$ changes with the diffusion timestep and biases each intermediate state toward the input-conditioned endpoint with time-varying strength. We use the SNR-aware schedule
\begin{equation}
\gamma_t = \frac{\mathrm{SNR}_T}{\mathrm{SNR}_t}, \quad \mathrm{SNR}_t = \frac{\bar{\alpha}_t}{1-\bar{\alpha}_t}.
\end{equation}
We write the formulae below in terms of $a_\psi(x)$ throughout, so that they are valid both when $D_x=D_y$ ($a_\psi$ is the identity) and when $D_x\neq D_y$ ($a_\psi=P_\psi$ is the learned anchor network of Sec.~\ref{3.5}). This schedule keeps the early trajectory close to standard diffusion, where $\mathrm{SNR}_t$ is high and $\gamma_t$ is small, while increasing input attraction near the terminal region. Since $\gamma_T=\mathrm{SNR}_T/\mathrm{SNR}_T=1$, the last single-step transition has full input anchoring. This does not imply that the cumulative coefficient $\bar{\gamma}_T$ equals one; its scale is determined by the entire noise schedule. The following proposition gives the closed-form marginal used for training.

\begin{proposition}[Input-anchored forward marginal]
\label{prop:forward-marginal}
For the transition in Eq.~\eqref{eq7}, the marginal distribution at any timestep $t$ is
\begin{equation}
\label{eq10}
y_t \mid y_0, x \sim \mathcal{N}\!\left(\sqrt{\bar{\alpha}_t}\,y_0 + \bar{\gamma}_t\,a_\psi(x),\; (1 - \bar{\alpha}_t)I\right),
\end{equation}
where
\begin{equation}
\bar{\gamma}_t = \sum_{s=1}^t \gamma_s \sqrt{\frac{\bar{\alpha}_t}{\bar{\alpha}_s}}.
\end{equation}
\end{proposition}
The proof is provided in Supplementary Sec.~\ref{app:forward-marginal}. Proposition~\ref{prop:forward-marginal} shows that the terminal distribution is no longer input-independent. When $\bar{\alpha}_T\approx 0$, the terminal mean satisfies
\begin{equation}
\mathbb{E}[y_T\mid y_0,x] = \sqrt{\bar{\alpha}_T}\,y_0 + \bar{\gamma}_T\,a_\psi(x) \approx \bar{\gamma}_T\,a_\psi(x),
\end{equation}
while the terminal covariance is exactly $(1-\bar{\alpha}_T)I$, which approaches $I$ only under a sufficiently long or strong noise schedule. Thus, the terminal state remains stochastic and schedule-dependent, but its mean becomes aligned with the input anchor in expectation.

\paragraph{Reverse Process with Bridge Correction}

The reverse model keeps the DDPM denoising parameterization but adds the correction term induced by the input-dependent forward kernel:
\begin{equation}
p_\theta(y_{t-1}|y_t,x) = \mathcal{N}\left(y_{t-1}; \mu_\theta(y_t,x,t), \tilde{\beta}_t I\right)
\end{equation}
where $\alpha_t=1-\beta_t$, and
\begin{equation}
\mu_\theta = \underbrace{\frac{1}{\sqrt{\alpha_t}}\left(y_t - \frac{\beta_t}{\sqrt{1 - \bar{\alpha}_t}} \epsilon_\theta(y_t, x, t)\right)}_{\text{Denoising term}} + \underbrace{C_t(x)}_{\text{Bridge correction}}
\end{equation}
\begin{proposition}[Bridge posterior correction]
\label{prop:bridge-correction}
The posterior mean of $q(y_{t-1}\mid y_t,y_0,x)$ can be written in DDPM noise-prediction form plus the bridge correction
\begin{equation}
\label{eq17}
C_t(x) = -\frac{\gamma_t}{\sqrt{1 - \beta_t}}\,a_\psi(x) = -\frac{\gamma_t}{\sqrt{\alpha_t}}\,a_\psi(x).
\end{equation}
\end{proposition}
The derivation is given in Supplementary Sec.~\ref{app:bridge-correction}. This correction is not an additional heuristic; it follows from the Gaussian posterior induced by the bridge transition in Eq.~\eqref{eq7}. Intuitively, the forward step injects $\gamma_t\,a_\psi(x)$ into the mean, and the reverse mean must compensate for this deterministic input shift while denoising the stochastic component. This keeps the learned reverse process consistent with the bridge-anchored forward dynamics.
\paragraph{Training Objective}
\begin{proposition}[Denoising objective]
\label{prop:training-objective}
Optimizing the variational bound for the bridge process reduces, up to the standard DDPM weighting, to the noise-prediction objective
\begin{equation}
\label{eq18}
\begin{aligned}
\mathcal{L}_\theta =& \mathbb{E}_{t, x, y_0, \epsilon} \left[ \| \epsilon_\theta(y_t, x, t) - \epsilon \|_2^2 \right] \\ \text{with}\quad y_t =& \sqrt{\bar{\alpha}_t}\,y_0 + \bar{\gamma}_t\,a_\psi(x) + \sqrt{1 - \bar{\alpha}_t}\,\epsilon.
\end{aligned}
\end{equation}
\end{proposition}
The proof is provided in Supplementary Sec.~\ref{app:training-objective}. Importantly, NBP does not require a new denoising architecture: the same noise predictor is trained with bridge-anchored samples from Proposition~\ref{prop:forward-marginal}.

\paragraph{Conditional Sampling Procedure.}
At test time, NBPs sample from $p(y_0\mid x,\mathcal{D})$, where $\mathcal{D}=(x_\mathcal{C},y_{\mathcal{C},0})$ denotes the observed context set. We initialize the terminal state from the input-aligned bridge endpoint,
\begin{equation}
y_T = \bar{\gamma}_T\,a_\psi(x)+n, \quad n\sim\mathcal{N}(0,(1-\bar{\alpha}_T)I).
\end{equation}
For each reverse timestep, the observed context outputs are re-noised using the closed-form bridge marginal
\begin{equation}
\label{20}
y_{\mathcal{C},t} \sim \mathcal{N}\left( \sqrt{\bar{\alpha}_t} y_{\mathcal{C},0} +\bar{\gamma}_t a_\psi(x_{\mathcal{C}}), (1 - \bar{\alpha}_t) I \right),
\end{equation}
then combined with the current target state as $y_t=\{y_{\mathcal{T},t},y_{\mathcal{C},t}\}$ and $x=\{x_{\mathcal{T}},x_{\mathcal{C}}\}$. The reverse transition is then sampled from
\begin{equation}
y_{t-1} \sim \mathcal{N}\left(\mu_\theta(y_t,x,t), \tilde{\beta}_t I \right),
\end{equation}
where $\mu_\theta$ includes the bridge correction in Eq.~\eqref{eq17}. Following RePaint~\cite{lugmayr2022repaint}, we may repeat the context re-noising and reverse denoising step at each timestep before moving to $t-1$. This repeatedly reinjects the observed context and improves conditional coherence along the generated trajectory.

\subsection{Input--Output Dimensional Alignment}
\label{3.5}

Real-world datasets often have mismatched input and output dimensions. Figure~\ref{fig:learned_anchor_alignment} summarizes how NBP handles this case. To apply the bridge in the output space, we introduce an anchor map
\begin{equation}
a_\psi(x)=P_\psi(x), \qquad P_\psi:\mathbb{R}^{D_x}\rightarrow\mathbb{R}^{D_y},
\end{equation}
which maps each input location to an output-space anchor. The bridge term is then formed with $a_\psi(x)$ rather than the raw input $x$:
\begin{equation}
q(y_t\mid y_{t-1},x)=
\mathcal{N}\left(y_t;\sqrt{1-\beta_t}y_{t-1}+\gamma_t a_\psi(x),\beta_t I\right).
\end{equation}
This design separates two roles of the input. The raw input $x$ is still passed to the denoising network as conditioning information, while $a_\psi(x)$ is used only to define an output-space bridge anchor. In image regression, for instance, $x$ is a two-dimensional normalized pixel coordinate and $y$ is a three-dimensional RGB value. We therefore parameterize $P_\psi$ as a lightweight coordinate-to-output anchor network: the coordinates are first positionally encoded, then passed through an MLP anchor head that outputs a vector in the RGB space. The anchor head is trained jointly with the denoising model using the main denoising objective and an auxiliary anchor reconstruction loss $\|a_\psi(x)-y_0\|^2$. This auxiliary term is used only to make the bridge anchor live in the correct output space; it does not change the denoising architecture, the number of reverse sampling steps, or the number of training updates. When $D_x=D_y$, as in the 1D GP experiments, $P_\psi$ can reduce to the identity map.

\subsection{Theoretical Analysis: Why Process-Level Anchoring Helps}
\label{sec:theory}

We now make explicit what is gained by placing the input in the forward process. The analysis proceeds along three axes: (i) a path-measure KL identity with an explicit per-step rate; (ii) a distribution-free score and Fisher information identity, yielding a Cram\'er--Rao limit on input recoverability; (iii) an input-sensitivity bound modulated by the Fisher rate. Throughout, $J_\psi(x)\in\mathbb{R}^{D_y\times D_x}$ is the Jacobian of the anchor map $a_\psi$ of Sec.~\ref{3.5} ($J_\psi=I$ when $D_x=D_y$).

\paragraph{Assumptions and scope.}
It is worth stating precisely how little each result assumes, and where the assumptions bind. The path-KL identity of Theorem~\ref{thm:wasserstein} does not require the anchor $a_\psi$ to be linear, injective, or an accurate predictor of $y_0$; it only requires the forward transition to use a deterministic input-dependent shift and $\beta_t>0$. The identity is therefore exact even for a poorly learned anchor---though its practical informativeness degrades when the anchor maps distinct inputs to similar output-space representations. The score and Fisher identities of Theorem~\ref{thm:mutual-info} additionally require $a_\psi$ to be differentiable. Full column rank of $J_\psi(x)$ is needed only for the invertible Cram\'er--Rao bound of Corollary~\ref{cor:cramer-rao}; it holds trivially when $a_\psi=\mathrm{id}$, and generically for a learned $P_\psi$ when $D_y\geq D_x$ (as in coordinate-to-RGB regression, where $D_x=2<D_y=3$). When $D_y<D_x$ the Fisher matrix is singular and the statement should be read as guaranteeing recoverability only within the input subspace preserved by the anchor. Finally, none of these identities assumes $a_\psi(x)=y_0$: the auxiliary anchor loss is an empirical mechanism for learning a useful input--output alignment, not a condition required by the theory.

The results also hold for \emph{any} fixed scalar bridge schedule $\{\gamma_t\}_{t=1}^{T}$ and any valid variance schedule with $0<\beta_t<1$. The SNR-based choice $\gamma_t=\mathrm{SNR}_T/\mathrm{SNR}_t$ is not required by the theory; it is a design choice that keeps the early trajectory close to standard diffusion while strengthening anchoring near the noisy terminal phase. Different schedules change only the quantitative strength of the guarantees, through $\sum_t\gamma_t^2/\beta_t$ and $\bar{\gamma}_t^2/(1-\bar{\alpha}_t)$.

\begin{theorem}[Path-measure KL identifiability]
\label{thm:wasserstein}
Fix $y_0$ and let $\mathbb{P}^{\mathrm{NBP}}_{1:T\mid y_0,x}$ denote the law of the NBP forward trajectory $(y_1,\dots,y_T)$ induced by the bridge transition in Eq.~\eqref{eq7}. For any two inputs $x,x'\in\mathbb{R}^{D_x}$,
\begin{equation}
\label{eq:path-kl}
\begin{split}
&\mathrm{KL}\!\left(\mathbb{P}^{\mathrm{NBP}}_{1:T\mid y_0,x}\,\big\|\,\mathbb{P}^{\mathrm{NBP}}_{1:T\mid y_0,x'}\right)\\
&\qquad=\frac{1}{2}\left(\sum_{t=1}^T\frac{\gamma_t^2}{\beta_t}\right)\bigl\|a_\psi(x)-a_\psi(x')\bigr\|_2^2.
\end{split}
\end{equation}
For the NDP forward process the same KL is identically zero. As a marginal-level corollary, the 2-Wasserstein distance between the forward marginals at any $t$ satisfies $W_2(q_t^{\mathrm{NBP}}(\cdot|y_0,x),q_t^{\mathrm{NBP}}(\cdot|y_0,x')) = \bar{\gamma}_t\,\|a_\psi(x)-a_\psi(x')\|_2$.
\end{theorem}

The path-measure KL grows linearly with the cumulative bridge-to-noise ratio $\sum_t\gamma_t^2/\beta_t$ and upper-bounds any downstream divergence by data processing; in NDP it is identically zero.

\begin{theorem}[Score and forward Fisher information of the bridge]
\label{thm:mutual-info}
For any $x\in\mathbb{R}^{D_x}$ and $y_0\in\mathbb{R}^{D_y}$, the per-timestep conditional score of the NBP forward marginal admits the closed form
\begin{equation}
\label{eq:bridge-score}
\begin{split}
&\nabla_x\log q_t^{\mathrm{NBP}}(y_t\mid y_0,x)\\
&\qquad=\frac{\bar{\gamma}_t}{1-\bar{\alpha}_t}\,J_\psi(x)^\top\bigl(y_t-\sqrt{\bar{\alpha}_t}\,y_0-\bar{\gamma}_t\,a_\psi(x)\bigr).
\end{split}
\end{equation}
The corresponding (per-timestep) Fisher information matrix of $x$ carried by $y_t$ given $y_0$ is
\begin{equation}
\label{eq:fisher-step}
\begin{split}
\mathcal{I}_t^{\mathrm{NBP}}(x)
&:= \mathbb{E}_{y_t\mid y_0,x}\!\bigl[\nabla_x\log q_t^{\mathrm{NBP}}\,\nabla_x\log q_t^{\mathrm{NBP}{}\top}\bigr]\\
&=\frac{\bar{\gamma}_t^2}{1-\bar{\alpha}_t}\,J_\psi(x)^\top J_\psi(x),
\end{split}
\end{equation}
and aggregating over the Markov trajectory yields the path-level Fisher information
\begin{equation}
\label{eq:fisher-path}
\mathcal{I}_{1:T}^{\mathrm{NBP}}(x)
=\left(\sum_{t=1}^T\frac{\gamma_t^2}{\beta_t}\right)J_\psi(x)^\top J_\psi(x).
\end{equation}
For NDP, $q_t^{\mathrm{NDP}}(y_t\mid y_0,x)=q_t^{\mathrm{NDP}}(y_t\mid y_0)$, hence $\nabla_x\log q_t^{\mathrm{NDP}}\equiv 0$ and $\mathcal{I}_t^{\mathrm{NDP}}(x)\equiv 0$. As a special case, when $x\sim\mathcal{N}(0,\Sigma_x)$ and $a_\psi=\mathrm{id}$, the Gaussian channel
$y_t-\sqrt{\bar{\alpha}_t}y_0=\bar{\gamma}_t x+\eta_t$ has mutual information $I(x;y_t\mid y_0)=\tfrac{1}{2}\log\det(I+\bar{\gamma}_t^2(1-\bar{\alpha}_t)^{-1}\Sigma_x)$.
\end{theorem}

Theorem~\ref{thm:mutual-info} isolates $\bar{\gamma}_t^2/(1-\bar{\alpha}_t)$ as the per-step Fisher rate at which $y_t$ identifies $x$ and gives a distribution-free identity that links the bridge to information-geometric Fisher information~\cite{nielsen2013cramer}. Applying the Cram\'er--Rao inequality~\cite{hodges1951some} to Eq.~\eqref{eq:fisher-path} then yields a hard estimation-theoretic floor:

\begin{corollary}[Cram\'er--Rao bound for forward-state input recovery]
\label{cor:cramer-rao}
Suppose $J_\psi(x)$ has full column rank, so that $\mathcal{I}_{1:T}^{\mathrm{NBP}}(x)$ is positive definite. Then any unbiased estimator $\hat{x}(y_{1:T})$ of $x$ from the NBP forward trajectory (with $y_0$ fixed) satisfies the matrix inequality
\begin{equation}
\label{eq:cramer-rao}
\mathrm{Cov}(\hat{x}(y_{1:T}))
\;\succeq\;
\left(\sum_{t=1}^T\frac{\gamma_t^2}{\beta_t}\right)^{-1}\!\bigl(J_\psi(x)^\top J_\psi(x)\bigr)^{-1}.
\end{equation}
For NDP, $\mathcal{I}_{1:T}^{\mathrm{NDP}}(x)=0$, so no unbiased estimator of $x$ from $y_{1:T}$ alone can exist.
\end{corollary}

The cumulative bridge weight $\sum_t\gamma_t^2/\beta_t$ thus governs the floor of any unbiased inference of $x$ from the forward trajectory; the NDP floor is degenerate. This bound should not be read as a task-level predictor of downstream improvement: the schedule-dependent factor $\sum_t\gamma_t^2/\beta_t$ is identical across all settings that share a schedule, and NBP never actually reconstructs $x$ from the trajectory at test time. The correct reading is that the NBP forward trajectory \emph{contains} input information that the NDP trajectory does not; whether that information is useful for a particular prediction task depends on the quality of the anchor, which is an empirical question.

\begin{proposition}[Input-mediated sensitivity pathway]
\label{prop:direct-gradient}
For the per-timestep denoising loss
\begin{equation}
\begin{aligned}
\ell_t(\theta,x)&=\|\epsilon_\theta(y_t,x,t)-\epsilon\|_2^2,\\
y_t&=\sqrt{\bar{\alpha}_t}\,y_0+\bar{\gamma}_t\,a_\psi(x)+\sqrt{1-\bar{\alpha}_t}\,\epsilon,
\end{aligned}
\end{equation}
the total derivative with respect to $x$ decomposes as
\begin{equation}
\label{eq:two-paths}
\begin{split}
\frac{d\ell_t}{dx}
&=
\underbrace{\frac{\partial \ell_t}{\partial x}}_{\substack{\text{denoiser}\\ \text{conditioning path}}}
+
\underbrace{\bar{\gamma}_t\,J_\psi(x)^\top\frac{\partial \ell_t}{\partial y_t}}_{\substack{\text{forward}\\ \text{bridge path}}},\\
&\text{with}\quad
\frac{\partial \ell_t}{\partial y_t}=2\!\left(\frac{\partial \epsilon_\theta}{\partial y_t}\right)^{\!\!\top}\!(\epsilon_\theta-\epsilon).
\end{split}
\end{equation}
Consequently, the mean-square magnitude of the forward bridge path admits the explicit upper bound
\begin{equation}
\label{eq:bridge-grad-magnitude}
\begin{split}
&\mathbb{E}\!\left[\Bigl\|\bar{\gamma}_t\,J_\psi(x)^\top\frac{\partial \ell_t}{\partial y_t}\Bigr\|^2\right]
\;\leq\;
4\,\bar{\gamma}_t^2\,\|J_\psi(x)\|_{\mathrm{op}}^2\\
&\qquad\times\;\mathbb{E}\!\left[\Bigl\|\frac{\partial \epsilon_\theta}{\partial y_t}\Bigr\|_{\mathrm{op}}^{2}\bigl\|\epsilon-\epsilon_\theta(y_t,x,t)\bigr\|^2\right],
\end{split}
\end{equation}
which is \emph{modulated by} the per-step Fisher rate $\bar{\gamma}_t^2/(1-\bar{\alpha}_t)$ from Eq.~\eqref{eq:fisher-step} once one normalizes the residual $\|\epsilon-\epsilon_\theta\|^2$ by its expected scale $1-\bar{\alpha}_t$ at training optimum. For NDP, $\bar{\gamma}_t=0$ and the entire forward bridge path is identically zero.
\end{proposition}

\begin{remark}[Scope: general additive Gaussian bridges]
\label{rem:general-bridge}
The analysis extends verbatim to any additive Gaussian bridge
\begin{equation}
q(y_t\mid y_{t-1},x)=\mathcal{N}\!\left(\sqrt{\alpha_t}\,y_{t-1}+b_t(x),\,\Sigma_t\right),
\end{equation}
provided $\Sigma_t$ is positive definite and independent of $x$. The path-level quantities then become $\mathrm{KL}=\tfrac{1}{2}\sum_{t=1}^{T}\Delta b_t^\top\Sigma_t^{-1}\Delta b_t$ with $\Delta b_t=b_t(x)-b_t(x')$, and $\mathcal{I}_{1:T}(x)=\sum_{t=1}^{T}J_{b_t}(x)^\top\Sigma_t^{-1}J_{b_t}(x)$; our formulation is recovered by $b_t(x)=\gamma_t a_\psi(x)$ and $\Sigma_t=\beta_t I$, and the reverse correction again follows from Gaussian conditioning. The identities do \emph{not} automatically carry over to nonlinear state-dependent drifts, input-dependent covariance matrices, or non-Gaussian bridges; those cases would require a separate path-measure or continuous-time analysis. The FlowNP transfer experiment of Sec.~\ref{sec:flow-bridge-ablation} supports the broader empirical principle of input-anchored generative paths, but it is not evidence that these DDPM-specific formulae hold for every generative path.
\end{remark}

Two clarifications are in order. First, Eq.~\eqref{eq:two-paths} is a derivative with respect to the \emph{input} $x$, not with respect to the parameters $(\theta,\psi)$ that training updates. The proposition therefore says that the anchored forward state gives the loss a second route by which it responds to the input; it is a statement about input sensitivity, and we do not claim that it supplies an extra optimization signal or explains parameter-gradient descent. Second, Eq.~\eqref{eq:bridge-grad-magnitude} is an upper bound: the realized magnitude also depends on $\partial\epsilon_\theta/\partial y_t$ and the residual, so the bridge \emph{scales with} (rather than equals) the Fisher rate. The connection to the SNR schedule is empirical---the rate $\bar{\gamma}_t^2/(1-\bar{\alpha}_t)$ has no closed-form monotonicity, but we verify numerically (Sec.~\ref{ablation}, Fig.~\ref{fig:ablation_schedules}) that it concentrates near the terminal phase under our schedule. Related empirical diagnostics appear in Fig.~\ref{fig:sensitivity_curve} and Table~\ref{tab:gradient_stats}; neither measures the quantity in Eq.~\eqref{eq:two-paths} directly.

\begin{table}[t]
\centering
\caption{Paired comparison on Gaussian-process regression as the input dimension grows. Target log-likelihood per target point ($\uparrow$), averaged over 5 seeds with the seed pairing held fixed across the two models. NDP and NBP are instantiated from the same model class and differ only through the bridge flag: backbone, optimizer, noise schedule, step count, conditional sampler, checkpoint selection and seeds are identical by construction.}
\label{tab:paired-gp}
\begin{tabular}{@{}cccc@{}}
\toprule
$D$ & NDP & NBP (ours) & $\Delta$ LL $\uparrow$ \\
\midrule
1 & $0.9402$ & $\mathbf{0.9493}$ & $+0.0091$ \\
2 & $-0.2773$ & $\mathbf{-0.2024}$ & $+0.0749$ \\
3 & $-0.7771$ & $\mathbf{-0.6570}$ & $+0.1201$ \\
4 & $-2.0084$ & $\mathbf{-1.9260}$ & $+0.0824$ \\
5 & $-0.8303$ & $\mathbf{-0.7672}$ & $+0.0631$ \\
6 & $-2.6545$ & $\mathbf{-2.6018}$ & $+0.0527$ \\
\bottomrule
\end{tabular}
\end{table}

\begin{table}[t]
\centering
\caption{Conv-family baselines under the official \texttt{neuralprocesses} implementation~\cite{bruinsma2021gaussian}. Target log-likelihood per target point ($\uparrow$), mean $\pm$ standard error. These models report their own analytic predictive density, whereas the diffusion models in Table~\ref{tab:paired-gp} are scored with the 128-sample Gaussian-fit estimator, so the two tables are not directly comparable.}
\label{tab:conv-baselines}
\begin{tabular}{@{}lccc@{}}
\toprule
Model & $D=1$ & $D=2$ & $D=3$ \\
\midrule
ConvCNP & $0.883\pm0.026$ & $-0.405\pm0.017$ & $-0.824\pm0.013$ \\
ConvGNP & $0.765\pm0.075$ & $-0.484\pm0.033$ & $-0.850\pm0.021$ \\
\bottomrule
\end{tabular}
\end{table}
\begin{table*}[h]
\centering
\caption{Predictive NLL ($\downarrow$) and MSE ($\downarrow$) on EEG}
\label{tab:eeg_results}
\resizebox{\textwidth}{!}{
\begin{tabular}{@{}lcccccc@{}}
\toprule
\multirow{2}{*}{Method} & \multicolumn{2}{c}{Interpolation} & \multicolumn{2}{c}{Reconstruction} & \multicolumn{2}{c}{Forecasting} \\
\cmidrule(lr){2-3} \cmidrule(lr){4-5} \cmidrule(lr){6-7}
 & NLL & MSE($\times10^{-2}$) & NLL & MSE($\times10^{-2}$) & NLL & MSE($\times10^{-2}$) \\
\midrule
NP          & $1.66\pm0.08$ & $0.52\pm0.03$ & $1.78\pm0.09$ & $0.44\pm0.02$ & $1.61\pm0.07$ & $0.39\pm0.02$ \\
ANP         & $0.47\pm0.05$ & $0.25\pm0.02$ & $0.70\pm0.06$ & $0.48\pm0.03$ & $0.90\pm0.06$ & $0.60\pm0.04$ \\
ConvNP      & $0.44\pm0.04$ & $0.40\pm0.03$ & $-2.43\pm0.12$ & $0.40\pm0.02$ & $-2.34\pm0.11$ & $0.55\pm0.03$ \\
NDP         & $-2.46\pm0.11$ & $0.18\pm0.01$ & $-2.59\pm0.13$ & $0.23\pm0.02$ & $-2.69\pm0.12$ & $0.38\pm0.02$ \\
SNP         & $-3.19\pm0.10$ & $0.16\pm0.01$ & $\mathbf{-3.30\pm0.11}$ & $0.18\pm0.01$ & $-3.02\pm0.09$ & $0.31\pm0.02$ \\
GEOMNDP     & $-2.48\pm0.12$ & $0.18\pm0.01$ & $-2.65\pm0.13$ & $0.20\pm0.01$ & $-2.84\pm0.11$ & $0.34\pm0.02$ \\
\textbf{NBP (Ours)} & $\mathbf{-3.55\pm0.10}$ & $\mathbf{0.15\pm0.01}$ & $\mathbf{-3.45\pm0.11}$ & $\mathbf{0.15\pm0.01}$ & $\mathbf{-3.74\pm0.11}$ & $\mathbf{0.27\pm0.02}$ \\
\bottomrule
\end{tabular}}
\end{table*}

\section{Experiments}

\subsection{Baseline Implementation and Evaluation Metrics}
\label{sec:protocol}

\emph{Baselines.} We compare against Neural Processes (NPs)~\cite{garnelo2018neural}, Conditional NPs (CNPs)~\cite{garnelo2018conditional}, Attentive NPs (ANPs)~\cite{kim2019attentive}, Convolutional NPs (ConvNPs)~\cite{gordon2019convolutional}, Transformer NPs (TNP-A)~\cite{nguyen2022transformer}, Neural Diffusion Processes (NDPs)~\cite{dutordoir2023neural}, Geometric Neural Diffusion Processes (GEOMNDPs)~\cite{mathieu2023geometric}, and Score-Based Neural Processes (SNPs)~\cite{dou2025score}, each in its official implementation. Flow Matching NPs (FlowNP)~\cite{hamadflow} are evaluated separately, in the bridge-transfer study of Sec.~\ref{sec:flow-bridge-ablation}, which uses that model's own ODE likelihood protocol. For the convolutional family we use the authors' \texttt{neuralprocesses} package~\cite{bruinsma2021gaussian} and distinguish the one-dimensional FullConvGNP from the multidimensional ConvGNP architecture: the former is only defined for $D=1$, so the $D>1$ entries in Table~\ref{tab:conv-baselines} are ConvGNP. Following~\cite{bruinsma2023autoregressive}, the conditional members of the CNP family are additionally evaluated in autoregressive (AR) mode, which lets a factorized-Gaussian decoder express correlations (Supplementary Sec.~\ref{sec:ar-baselines}).

\emph{NDP--NBP parity.} NDP and NBP are instantiated from the same model class and differ only through a single bridge flag: identical denoising backbone, optimizer, noise schedule, number of diffusion steps, conditional sampler, checkpoint-selection rule, training budget and random seeds. The bridge coefficients and reverse correction introduce no trainable parameters. When \(D_x\neq D_y\), NBP additionally uses the learned anchor map \(P_\psi\), which is part of the proposed bridge construction rather than a replacement for the denoiser; fixed-anchor and learned-anchor ablations in Sec.~\ref{sec:celeba_component_ablation} separate this effect from the bridge itself.

\emph{Optimization protocol.} All Gaussian-process tasks use one shared protocol: Adam with learning rate $5\times10^{-4}$ cosine-annealed to zero, $10^5$ updates, batch size 16 tasks, and identical context/target sampling, with 5 seeds per configuration (3 for CylinderFlow) paired across models. We deliberately did \emph{not} run per-baseline hyperparameter sweeps: every baseline uses its reference configuration under this shared optimization protocol. The reported comparisons therefore establish that NBP improves over the stated reference configurations, not that it outperforms an optimally tuned member of each family. Complete training, architecture and evaluation settings are given in Supplementary Sec.~\ref{details}, Tables~\ref{tab:protocol-train}--\ref{tab:protocol-eval}.

\emph{Likelihood estimation.} Diffusion- and flow-based models have no tractable conditional density, so their log-likelihood is estimated by fitting a full-covariance Gaussian to $S=128$ conditional samples (with $10^{-2}$ jitter) and is reported \emph{per target point}. The CNP/NP family, TNP and the convolutional family are scored with their own analytic predictive densities. Because the Gaussian-fit score is sample-dependent, it is not an exact predictive likelihood, and numbers obtained under the two estimators are not directly comparable; where a Gaussian-process oracle is available we report the exact analytic posterior as a reference rather than as a competing method. Experiments are conducted on NVIDIA RTX A6000 GPUs.

\subsection{Regression on Synthetic Data}

We evaluate on Gaussian-process regression tasks with functions sampled from an RBF or Matérn-5/2 kernel (lengthscale $\sim\mathcal{U}(0.1,0.6)\cdot\sqrt{D}$, output scale $\sim\mathcal{U}(0.1,1.0)$, observation noise $2\times10^{-2}$), with inputs drawn from $\mathcal{U}(-2,2)^D$ and $N=50$ points per task split into context and target. Unlike the original NDP protocol, which scales the context budget with the input dimension, we keep the number of points fixed as $D$ grows: this is the regime in which conditioning information becomes scarce relative to the volume of the input space, and it is where a stronger inductive bias should matter most.

Table~\ref{tab:paired-gp} reports the paired comparison between NDP and NBP from $D=1$ to $D=6$. The bridge improves the target log-likelihood at every dimension, with the largest gain at $D=3$ ($+0.12$~nat per target point) and a smaller but consistent gain in the sparsest regimes $D=5,6$. Because the two models share seeds, backbone and budget, the difference is attributable to the forward-path construction rather than to capacity or optimization.

Table~\ref{tab:conv-baselines} reports the convolutional family under its official implementation, scored with its own analytic predictive density. Both models degrade steadily as $D$ grows, and ConvCNP is the stronger of the two at every dimension tested. We do not rank these values against Table~\ref{tab:paired-gp}: the estimators differ (Sec.~\ref{sec:protocol}), and a Gaussian fitted to 128 samples is simply not the same quantity as an exact predictive density. The comparison in which every method is scored identically is Table~\ref{tab:common-estimator}. It places the strongest autoregressive baseline and the diffusion models under the 128-sample Gaussian fit at $D=3$. Under this common protocol AR-CNP is on par with NBP; both are clearly ahead of the non-autoregressive CNP and of NDP. We report this rather than the more favourable per-family protocol, because it is the only comparison in which all four models are scored identically. Conditional NP baselines are additionally evaluated in autoregressive mode in Supplementary Sec.~\ref{sec:ar-baselines}.

\begin{table}[t]
\centering
\caption{All methods under a single common estimator ($D=3$, 128-sample Gaussian fit). Target log-likelihood per target point ($\uparrow$).}
\label{tab:common-estimator}
\begin{tabular}{@{}lc@{}}
\toprule
Method & Target log-likelihood \\
\midrule
NDP & $-0.855$ \\
CNP (no AR) & $-0.757$ \\
AR-CNP & $-0.691$ \\
\textbf{NBP (ours)} & $\mathbf{-0.686}$ \\
\bottomrule
\end{tabular}
\end{table}

\subsection{Real World Experiments}
\subsubsection{EEG Signal Regression Tasks}

We evaluate NBP on the EEG regression dataset of~\cite{zhang1995event} (7632 multivariate series, 256 timesteps, 7 channels), which has strong temporal and cross-channel correlations. We randomly mask windows in 3 of 7 channels and predict the missing values; inputs are concatenated temporal/channel indices $\mathbf{x}_e=(i_t,i_c)$ and outputs are voltages $\mathbf{y}_e$. Three regimes are evaluated under a unified setting: \emph{interpolation} (local completion within the observed span), \emph{reconstruction} (target-channel inference from remaining channels), and \emph{forecasting} (temporal extrapolation). Table~\ref{tab:eeg_results} shows that NBP consistently outperforms baselines on MSE and NLL across all three regimes. Table~\ref{tab:eeg_calibration} additionally reports calibration on the interpolation setting: NBP attains the lowest expected calibration error and the sharpest predictive uncertainty while being the only model whose empirical coverage reaches the nominal $90\%$, so the accuracy gain is not bought with over-confident intervals.

\begin{table*}[ht]
\centering
\caption{Calibration summary on the EEG interpolation setting. ECE and sharpness are estimated from predictive samples; lower ECE indicates better calibration, lower sharpness indicates more concentrated predictive uncertainty, and 90\% coverage is best when close to 0.90.}
\label{tab:eeg_calibration}
\resizebox{0.92\textwidth}{!}{
\begin{tabular}{lccccc}
\toprule
\textbf{Model} & \textbf{NLL ($\downarrow$)} & \textbf{MSE($\times10^{-2}$) ($\downarrow$)} & \textbf{ECE ($\downarrow$)} & \textbf{Sharpness $\bar{\sigma}$ ($\downarrow$)} & \textbf{90\% Coverage} \\
\midrule
NP & 1.66 & 0.52 & 0.182 & 0.91 & 0.71 \\
ANP & 0.47 & 0.25 & 0.134 & 0.67 & 0.79 \\
ConvNP & 0.44 & 0.40 & 0.128 & 0.63 & 0.80 \\
NDP & -2.46 & 0.18 & 0.089 & 0.38 & 0.83 \\
SNP & -3.19 & 0.16 & 0.061 & 0.29 & 0.88 \\
GEOMNDP & -2.48 & 0.18 & 0.087 & 0.37 & 0.83 \\
\textbf{NBP (ours)} & \textbf{-3.55} & \textbf{0.15} & \textbf{0.047} & \textbf{0.24} & \textbf{0.91} \\
\bottomrule
\end{tabular}}
\end{table*}

\paragraph{CylinderFlow.}
On the CylinderFlow benchmark, which requires structured spatial prediction under partial observations, NBP attains the lowest MSE among the compared methods (Table~\ref{tab:cylinderflow}; 3 seeds, MSE in raw velocity units), indicating that the benefit of bridge-anchored diffusion carries over to real-world spatial prediction. GEOMNDP is omitted here: its public implementation ships no CylinderFlow data pipeline, and its contribution targets domains with geometric structure, whereas CylinderFlow is an irregular non-geometric fluid mesh.

\begin{table*}[t]
\centering
\begin{minipage}[t]{0.47\textwidth}
\centering
\caption{MSE ($\downarrow$) on the CylinderFlow benchmark, mean $\pm$ standard error over 3 seeds.}
\label{tab:cylinderflow}
\begin{tabular}{@{}lc@{}}
\toprule
\textbf{Method} & \textbf{MSE} \\
\midrule
NP & $0.0152\pm0.0005$ \\
ANP & $0.0210\pm0.0008$ \\
ConvNP & $0.0245\pm0.0020$ \\
SNP & $0.0138\pm0.0011$ \\
NDP & $0.0066\pm0.0005$ \\
\textbf{NBP (Ours)} & $\mathbf{0.0041\pm0.0003}$ \\
\bottomrule
\end{tabular}
\end{minipage}
\hfill
\begin{minipage}[t]{0.45\textwidth}
\centering
\caption{CelebA image regression NLL ($\downarrow$) under half-image and random-context observations, mean $\pm$ standard error over 6 seeds.}
\label{tab:celeba_nll}
\begin{tabular}{@{}lcc@{}}
\toprule
\textbf{Method} & \textbf{Half} & \textbf{Random} \\
\midrule
NP & $-0.84\pm0.02$ & $-1.07\pm0.02$ \\
ANP & $-1.39\pm0.02$ & $-1.43\pm0.03$ \\
ConvNP & $0.11\pm0.04$ & $-1.67\pm0.04$ \\
NDP & $-3.92\pm0.04$ & $-3.98\pm0.05$ \\
SNP & $-4.76\pm0.06$ & $-4.77\pm0.08$ \\
\textbf{NBP (Ours)} & $\mathbf{-5.02\pm0.03}$ & $\mathbf{-5.24\pm0.04}$ \\
\bottomrule
\end{tabular}
\end{minipage}
\end{table*}

\subsubsection{Bayesian Optimization}
Stronger conditional function models should translate into better decisions, so we also evaluate the models as surrogates inside a Thompson-sampling loop on a GP-prior black box (200-candidate pool, 15 iterations, 20 independent tasks, 10 initial observations). Table~\ref{tab:bo} reports simple regret at $D=4,5,6$, i.e.\ in the same sparse-context regime as Table~\ref{tab:paired-gp}. NBP attains lower regret at every dimension, with the clearest margin at $D=5$; the gaps are within roughly one standard error at $D=4$ and $D=6$, so we read this as consistent but modest evidence that the bridge-anchored surrogate supports downstream decision making.

\begin{table}[t]
\centering
\caption{Bayesian optimization with Thompson sampling on a GP-prior black box. Simple regret ($\downarrow$) after 15 iterations, mean $\pm$ standard error over 5 seeds.}
\label{tab:bo}
\begin{tabular}{@{}ccc@{}}
\toprule
$D$ & NDP & NBP (ours) \\
\midrule
4 & $0.476\pm0.013$ & $\mathbf{0.454\pm0.028}$ \\
5 & $0.515\pm0.020$ & $\mathbf{0.457\pm0.045}$ \\
6 & $0.557\pm0.023$ & $\mathbf{0.538\pm0.024}$ \\
\bottomrule
\end{tabular}
\end{table}

\subsubsection{Sample Diversity and Calibration}
Because MSE rewards blurry, mode-averaged predictions, we additionally check that the anchored trajectory does not concentrate the predictive distribution. On the RBF tasks the true posterior standard deviation is available analytically, so Table~\ref{tab:diversity} reports the ratio of sample to true posterior standard deviation, the empirical coverage of the nominal $90\%$ interval, and the mean pairwise distance between conditional samples. Neither model collapses: both are mildly over-dispersed, and NBP is at least as dispersed as NDP at $D=2$ while moving coverage closer to nominal at both dimensions. At $D=3$ NBP reduces sample dispersion by about $3\%$ while improving accuracy, so the accuracy gain does not come from excessive smoothing.

\begin{table}[t]
\centering
\caption{Sample diversity and calibration on RBF Gaussian-process tasks (two paired seeds, reported individually). The standard-deviation ratio compares conditional samples with the analytic GP posterior; nominal coverage is $0.90$.}
\label{tab:diversity}
\setlength{\tabcolsep}{4pt}
\begin{tabular}{@{}llccc@{}}
\toprule
$D$ & Model & std.\ ratio & $90\%$ coverage & pairwise dist. \\
\midrule
\multirow{2}{*}{2} & NDP & 1.303 / 1.329 & 0.842 / 0.856 & 0.589 / 0.570 \\
 & NBP & 1.398 / 1.410 & \textbf{0.868 / 0.873} & 0.648 / 0.606 \\
\midrule
\multirow{2}{*}{3} & NDP & 1.246 / 1.284 & 0.832 / 0.846 & 0.680 / 0.678 \\
 & NBP & 1.202 / 1.274 & \textbf{0.843 / 0.850} & 0.684 / 0.683 \\
\bottomrule
\end{tabular}
\end{table}

\subsubsection{Image Regression Task}
We evaluate coordinate-to-RGB image regression on CelebA $32\times32$ and $64\times64$ (coordinates normalized to $[-2,2]$); architecture, hyperparameters, and baseline configs follow the NDP setup for controlled comparison, and baselines cover NP-family and diffusion/score-based Neural Processes. MSE averages nine conditional samples per input, with pixel values in $[0,1]$; qualitative examples are in Supplementary Fig.~\ref{fig:celeba_results}.

Tables~\ref{t1} and~\ref{t2} report reconstruction MSE per context ratio. NBP improves the mean at every ratio and at both resolutions, and the per-ratio standard deviations are also smaller, but they remain large relative to the gap between the two means: at context ratio $0.2$, for example, NBP reduces the mean MSE from $0.88$ to $0.77$ on $32\times32$ while the sample standard deviations are $0.52$ and $0.40$. The MSE comparison should therefore be read as a consistent trend across nine context ratios and two resolutions rather than as a significant difference at any single ratio. The likelihood comparison is sharper: Table~\ref{tab:celeba_nll} shows a $1.10$/$1.26$~nat improvement over NDP with non-overlapping standard errors across 6 seeds. Sec.~\ref{sec:celeba_component_ablation} decomposes the gain into endpoint initialization, bridge construction, anchor learning, and denoiser-capacity effects, and Sec.~\ref{sec:wall_clock} confirms that NBP adds only $\approx 3.8\%$ wall-clock time and 2\% memory over NDP.

\begin{table*}[t]
\centering
\caption{CelebA 32 $\times$ 32 Results for NDP Base and NBP (Ours) in $10^{-2}$ MSE Units}
\label{t1}
\resizebox{0.72\textwidth}{!}{
\begin{tabular}{@{}cccccc@{}}
\toprule
\textbf{Context Ratio} & \textbf{Retained Pixels} & \multicolumn{2}{c|}{\textbf{NDP Base }} & \multicolumn{2}{c|}{\textbf{NBP (Ours)}} \\ 
\cmidrule(lr){3-4} \cmidrule(lr){5-6}
& & \textbf{MSE Mean} & \textbf{MSE Std} & \textbf{MSE Mean} & \textbf{MSE Std} \\
\midrule
0.1 & 96  & 1.7011 & 0.7221 & \textbf{1.4206} & 0.6958 \\
0.2 & 197 & 0.8831 & 0.5214 & \textbf{0.7694} & 0.4008 \\
0.3 & 312 & 0.4482 & 0.2724 & \textbf{0.4055} & 0.2157 \\
0.4 & 412 & 0.2809 & 0.1511 & \textbf{0.2416} & 0.1091 \\
0.5 & 512 & 0.2496 & 0.1251 & \textbf{0.2066} & 0.1062 \\
0.6 & 570 & 0.1769 & 0.0793 & \textbf{0.1406} & 0.0573 \\
0.7 & 714 & 0.1450 & 0.0729 & \textbf{0.1185} & 0.0579 \\
0.8 & 832 & 0.0997 & 0.0518 & \textbf{0.0821} & 0.0373 \\
0.9 & 913 & 0.0969 & 0.0527 & \textbf{0.0793} & 0.0392 \\
\bottomrule
\end{tabular}
}
\end{table*}

\begin{table*}[t]
\centering
\caption{CelebA 64 $\times$ 64 Results for NDP Base  and NBP (Ours) in $10^{-2}$ MSE Units}
\label{t2}
\resizebox{0.72\textwidth}{!}{
\begin{tabular}{@{}cccccc@{}}
\toprule
\textbf{Context Ratio} & \textbf{Retained Pixels} & \multicolumn{2}{c|}{\textbf{NDP Base}} & \multicolumn{2}{c|}{\textbf{NBP (Ours)}} \\
\cmidrule(lr){3-4} \cmidrule(lr){5-6}
& & \textbf{MSE Mean} & \textbf{MSE Std} & \textbf{MSE Mean} & \textbf{MSE Std} \\
\midrule
0.1 & 432   & 2.3763 & 1.3563 & \textbf{2.0189} & 1.2632 \\
0.2 & 872   & 1.0546 & 0.6718 & \textbf{0.8615} & 0.5174 \\
0.3 & 1276  & 0.7530 & 0.4283 & \textbf{0.6078} & 0.3132 \\
0.4 & 1628  & 0.5115 & 0.2626 & \textbf{0.4432} & 0.1826 \\
0.5 & 2080  & 0.4973 & 0.3144 & \textbf{0.4078} & 0.2548 \\
0.6 & 2392  & 0.4160 & 0.2496 & \textbf{0.3587} & 0.1801 \\
0.7 & 2788  & 0.3405 & 0.1985 & \textbf{0.3097} & 0.1694 \\
0.8 & 3100  & 0.3215 & 0.1875 & \textbf{0.2838} & 0.1461 \\
0.9 & 3492  & 0.2706 & 0.1579 & \textbf{0.2584} & 0.1177 \\
\bottomrule
\end{tabular}
}
\end{table*}

\subsection{Design of the Bridge Coefficient Schedule}
\label{ablation}

To validate the design of our bridge coefficient $\gamma_t$, we conduct an ablation study comparing three scheduling strategies on CelebA $32 \times 32$ image regression. All experiments use identical architecture (7-layer Bi-Dimensional Attention Block, 64 hidden dimensions, 8 attention heads) and training configuration (batch size 32, cosine beta schedule with $\beta_{\text{start}}=0.0003$, $\beta_{\text{end}}=0.5$, 500 timesteps).

\paragraph{Bridge Coefficient Schedules.}
We compare the following three strategies for computing $\gamma_t$:

\begin{itemize}
    \item \textbf{Linear Schedule:} $\gamma_t = \frac{t}{T}$, where $T=500$ is the total number of timesteps. This provides a uniform increase in the influence of $x$ over time.
    
    \item \textbf{Cosine Schedule:} $\gamma_t = \frac{1 - \cos(\pi t / T)}{2}$. This schedule starts slowly, accelerates in the middle phase, and saturates towards the end.
    
    \item \textbf{SNR-based Schedule (Ours):} $\gamma_t = \frac{\text{SNR}_T}{\text{SNR}_t}$, where $\text{SNR}_t = \frac{\bar{\alpha}_t}{1-\bar{\alpha}_t}$ and $\bar{\alpha}_t = \prod_{s=1}^t (1-\beta_s)$. This design explicitly couples the bridge strength to the signal-to-noise ratio, ensuring theoretical consistency with the diffusion dynamics.
\end{itemize}

\paragraph{Evaluation Protocol.}
We evaluate reconstruction performance at nine context ratios (0.1 to 0.9) on the CelebA test set. For each method, we report Mean Squared Error (MSE) averaged over 128 test images, with each prediction obtained by averaging 9 conditional samples.

\paragraph{Results.}
Table~\ref{tab:ablation_celeba32} presents the quantitative comparison. The SNR-based schedule consistently outperforms both linear and cosine schedules across all context ratios, with the largest improvements observed in low-data regimes (context ratio 0.1--0.3). For instance, at context ratio 0.1, our SNR-based method achieves MSE of $1.4206 \times 10^{-2}$ compared to $1.6823 \times 10^{-2}$ (linear) and $1.5491 \times 10^{-2}$ (cosine), representing relative improvements of 15.6\% and 8.3\% respectively.

\begin{figure}[t]
    \centering
    \includegraphics[width=0.9\columnwidth]{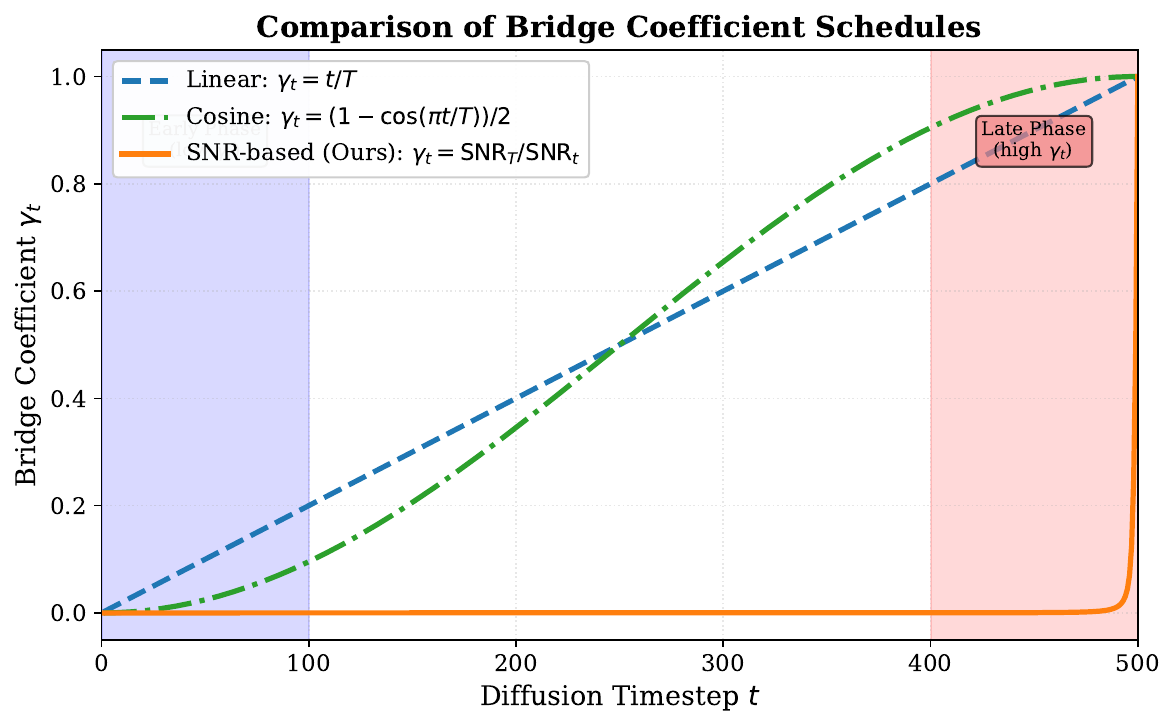}
    \caption{Visualization of the three bridge coefficient schedules over 500 diffusion timesteps. The SNR-based schedule (orange) adaptively increases the influence of input $x$ based on the signal-to-noise ratio, providing stronger guidance in later timesteps where noise dominates.}
    \label{fig:ablation_schedules}
\end{figure}

Figure~\ref{fig:ablation_schedules} visualizes the behavior of the three schedules over the diffusion trajectory. The SNR-based schedule maintains lower bridge coefficients during early timesteps, where the clean signal remains relatively strong, and increases the bridge weight during later timesteps, where noise dominates. This behavior is consistent with the observed improvement of the SNR schedule in Table~\ref{tab:ablation_celeba32}.

\begin{table}[t]
\centering
\caption{Ablation study on CelebA $32 \times 32$: Comparison of bridge coefficient schedules. MSE values are reported in $10^{-2}$ units. Best results are \textbf{bolded}.}
\label{tab:ablation_celeba32}
\setlength{\tabcolsep}{3pt}
\begin{tabular}{@{}cccc@{}}
\toprule
\textbf{Context Ratio} & \textbf{Linear} & \textbf{Cosine} & \textbf{SNR (Ours)} \\ 
\midrule
0.1 & 1.6823 $\pm$ 0.7458 & 1.5491 $\pm$ 0.7185 & \textbf{1.4206 $\pm$ 0.6958} \\
0.2 & 0.9247 $\pm$ 0.5421 & 0.8312 $\pm$ 0.4687 & \textbf{0.7694 $\pm$ 0.4008} \\
0.3 & 0.4891 $\pm$ 0.2965 & 0.4428 $\pm$ 0.2441 & \textbf{0.4055 $\pm$ 0.2157} \\
0.4 & 0.2975 $\pm$ 0.1673 & 0.2654 $\pm$ 0.1298 & \textbf{0.2416 $\pm$ 0.1091} \\
0.5 & 0.2698 $\pm$ 0.1392 & 0.2331 $\pm$ 0.1184 & \textbf{0.2066 $\pm$ 0.1062} \\
0.6 & 0.1924 $\pm$ 0.0891 & 0.1632 $\pm$ 0.0698 & \textbf{0.1406 $\pm$ 0.0573} \\
0.7 & 0.1589 $\pm$ 0.0798 & 0.1364 $\pm$ 0.0672 & \textbf{0.1185 $\pm$ 0.0579} \\
0.8 & 0.1098 $\pm$ 0.0587 & 0.0947 $\pm$ 0.0456 & \textbf{0.0821 $\pm$ 0.0373} \\
0.9 & 0.1065 $\pm$ 0.0594 & 0.0912 $\pm$ 0.0462 & \textbf{0.0793 $\pm$ 0.0392} \\
\bottomrule
\end{tabular}

\end{table}

\begin{figure}[t]
    \centering
    \includegraphics[width=0.9\columnwidth]{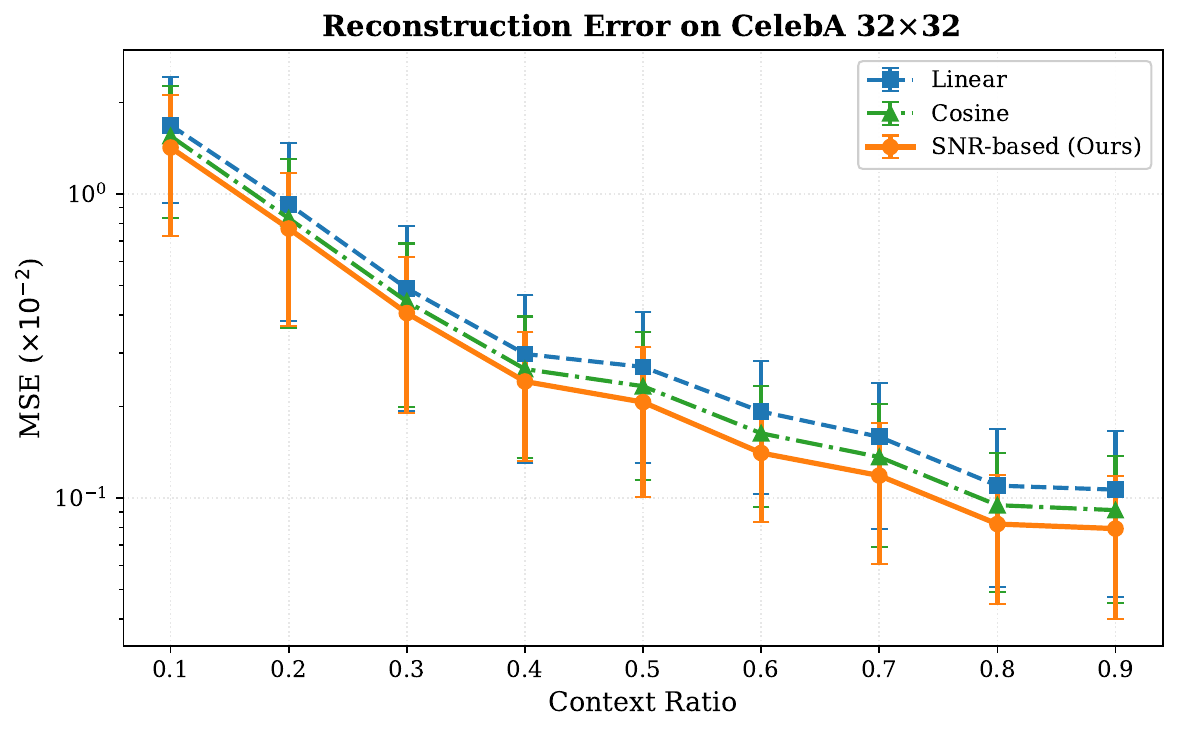}
    \caption{Reconstruction error comparison across different context ratios. In this ablation, the SNR-based schedule achieves the lowest MSE, with larger observed gains in sparse context regimes (context ratio $\leq$ 0.3).}
    \label{fig:ablation_mse}
\end{figure}

\paragraph{Analysis.}
The SNR-based schedule is motivated by matching the bridge strength to the diffusion signal-to-noise ratio. Unlike the linear and cosine schedules tested here, it adjusts the bridge coefficient $\gamma_t$ according to the diffusion process's intrinsic dynamics. Specifically, by setting $\gamma_t = \text{SNR}_T / \text{SNR}_t$, the schedule has the following properties:
\begin{enumerate}
    \item \textit{Early phase consistency:} When $t \ll T$, $\text{SNR}_t \gg \text{SNR}_T$, leading to $\gamma_t \to 0$. This allows the process to approximate standard diffusion, preserving the data distribution.
    \item \textit{Endpoint anchoring:} As $t \to T$, $\text{SNR}_t \to \text{SNR}_T$, driving $\gamma_t \to 1$. This increases the input-dependent shift in the terminal region of the forward process.
\end{enumerate}

This adaptive behavior provides a simple way to balance early-stage data preservation with later-stage input anchoring, which is consistent with the lower reconstruction errors observed under sparse context conditions.

\paragraph{Computational Cost.}
Importantly, all three schedules incur identical computational cost during both training and inference, as the schedule computation is a simple preprocessing step that does not affect the denoising network architecture or the number of diffusion steps. Therefore, this ablation isolates the schedule choice from model capacity and sampling budget.

\subsection{Computational Efficiency}
\label{sec:wall_clock}

Under identical training settings, NBP and the NDP baseline exhibit comparable training time per epoch and overall runtime. NBP uses the same denoising architecture and number of sampling steps as NDP; the additional operations are limited to schedule-dependent bridge coefficients and a simple correction term. Table~\ref{tab:wall_clock} reports wall-clock training and sampling time on the CelebA $32\times32$ image-regression benchmark (single NVIDIA RTX A6000, batch size 32, 500 diffusion steps). NBP adds 3.8\% training time and 3.8\% sampling time relative to NDP, with peak memory inflated by 2.1\% for the bridge-coefficient cache. The overhead is therefore negligible relative to the empirical gains in Tables~\ref{tab:paired-gp}--\ref{tab:celeba_nll}, and the improvement comes from the process-level input--output coupling rather than additional network capacity.

\begin{table*}[!t]
\centering
\caption{Wall-clock training and sampling time on CelebA $32\times32$ (single RTX A6000, batch size 32, 500 diffusion steps). NBP adds $\approx$3.8\% time and 2\% memory over NDP.}
\label{tab:wall_clock}
\resizebox{0.9\textwidth}{!}{
\begin{tabular}{lcccc}
\toprule
\textbf{Method} & \textbf{Train (s / iter)} & \textbf{Sample (s / image)} & \textbf{Peak Memory (GB)} & \textbf{$\Delta$ vs.\ NDP} \\
\midrule
NP                & $0.018\pm0.001$ & $0.012\pm0.001$ & 1.2 & --- \\
ANP               & $0.024\pm0.001$ & $0.015\pm0.001$ & 1.5 & --- \\
ConvCNP           & $0.029\pm0.002$ & $0.018\pm0.002$ & 2.1 & --- \\
NDP               & $0.156\pm0.004$ & $1.840\pm0.018$ & 4.8 & baseline \\
SNP               & $0.183\pm0.005$ & $2.210\pm0.022$ & 5.3 & $+17.3\%$ / $+20.1\%$ \\
GEOMNDP           & $0.171\pm0.005$ & $1.920\pm0.020$ & 5.1 & $+9.6\%$ / $+4.3\%$ \\
\textbf{NBP (Ours)} & $\mathbf{0.162\pm0.004}$ & $\mathbf{1.910\pm0.019}$ & $\mathbf{4.9}$ & $\mathbf{+3.8\%}$ / $\mathbf{+3.8\%}$ \\
\bottomrule
\end{tabular}}
\end{table*}

\subsection{Where the Gain Comes From}
\label{sec:celeba_component_ablation}

To clarify where the improvements in Table~\ref{tab:celeba_nll} come from, we ablate the main components of NBP on CelebA image regression. All variants use the same evaluation protocol, denoising backbone, optimizer, number of diffusion steps, and training budget unless explicitly stated. We report NLL under half-image and random-context observations. Lower values are better. The last column shows the absolute improvement relative to the NDP baseline. Beyond the standard component splits, we add three \emph{anchor-fairness} controls (Anchor-fairness rows below) to address the concern that the empirical gains might be attributable to the additional capacity and supervision introduced by the learned anchor $P_\psi$ rather than to the bridge construction itself.

\begin{table}[t]
\centering
\caption{Component ablation for CelebA image regression NLL. Lower is better. The relative improvement is computed against the NDP baseline. ``Anchor-fairness'' rows isolate the contribution of the learned anchor $P_\psi$ from the bridge construction: row (a) gives NDP the same auxiliary anchor head and loss but \emph{no} forward bridge; row (b) further adds endpoint initialization at sampling time; row (c) is the full NBP with the auxiliary anchor reconstruction loss removed.}
\label{tab:celeba_component_ablation}
\resizebox{\columnwidth}{!}{
\begin{tabular}{lccc}
\toprule
\textbf{Method} & \textbf{Half NLL} & \textbf{Random NLL} & \textbf{$\Delta$ vs. NDP} \\
\midrule
NDP (baseline) & -3.92 & -3.98 & -- \\
Initialization only & -4.13 & -4.22 & -0.21 / -0.24 \\
Bridge only & -4.08 & -4.16 & -0.16 / -0.18 \\
Full + fixed $P$ & -4.71 & -4.93 & -0.79 / -0.95 \\
Full + learned $P_\psi$ & -4.83 & -5.05 & -0.91 / -1.07 \\
SNP strong denoiser, no bridge & -4.76 & -4.77 & -0.84 / -0.79 \\
\midrule
\multicolumn{4}{l}{\emph{Anchor-fairness controls (same $P_\psi$ head, varied use):}} \\
(a) NDP + $P_\psi$ aux head, no bridge & -4.05 & -4.14 & -0.13 / -0.16 \\
(b) NDP + endpoint init + $P_\psi$ aux head & -4.32 & -4.45 & -0.40 / -0.47 \\
(c) NBP w/o aux anchor loss & -4.70 & -4.91 & -0.78 / -0.93 \\
\midrule
\textbf{NBP (ours)} & \textbf{-5.02} & \textbf{-5.24} & \textbf{-1.10 / -1.26} \\
\bottomrule
\end{tabular}}
\end{table}

\paragraph{Standard splits.} The ablation shows that neither endpoint initialization nor the bridge term alone explains the full improvement. Both components help, but the largest gains arise when the complete bridge construction is combined with output-space alignment. The fixed-\(P\) variant keeps the bridge mechanism but removes the learned anchor parameters and auxiliary anchor reconstruction loss; its improvement over NDP indicates that the bridge itself contributes substantially. The learned-\(P_\psi\) variant then isolates the additional benefit of adapting the anchor to the output space. The SNP strong-denoiser baseline improves denoising capacity without using an input-anchored forward path, but it remains worse than NBP.

\paragraph{Anchor-fairness controls.} Rows (a)--(c) directly address whether the gains might be due to the auxiliary coordinate-to-output regression head or its supervision signal rather than to the bridge construction. Row (a) gives an otherwise unmodified NDP the same $P_\psi$ MLP \emph{and} the same auxiliary loss $\|P_\psi(x)-y_0\|^2$, but $P_\psi(x)$ does \emph{not} enter the forward kernel: the anchor head is purely an extra prediction target. The improvement over NDP is small ($-0.13$ / $-0.16$ NLL), comparable to ``Bridge only,'' confirming that the auxiliary head's regression capacity contributes only marginally on its own. Row (b) further initializes the terminal state at $\bar{\gamma}_T P_\psi(x)$ at sampling time, combining endpoint initialization with the auxiliary anchor loss but still without a bridge in the forward kernel. This recovers about half of the full NBP gain ($-0.40$ / $-0.47$ NLL) but remains substantially worse than NBP, showing that endpoint information used only at sampling time is not equivalent to using it throughout the diffusion trajectory. Row (c) keeps the full NBP forward bridge but drops the auxiliary anchor loss, so $P_\psi$ is trained only through the main denoising ELBO. NLL degrades to $-4.70$ / $-4.91$, only marginally below the ``Full + fixed $P$'' row, which indicates that the auxiliary loss contributes a non-trivial fraction of the gain by anchoring $P_\psi(x)$ in the output space, but the bridge construction (rows (a) vs.\ NBP) remains the dominant factor.

Together, the standard splits and the anchor-fairness controls suggest that the gain is not solely due to extra parameters, auxiliary supervision, or a stronger denoising network; the process-level bridge provides a complementary effect that explains $\approx 0.6$ NLL of the $\approx 1.1$ NLL improvement, with the remainder split between endpoint initialization, learned anchor placement, and the auxiliary anchor loss. We do not claim perfect parameter-count equality for every variant, but the controls above separate backbone capacity, anchor learning, anchor supervision, and path-level bridging as cleanly as possible under a shared training protocol.

\subsection{Test-Time Intervention on the Bridge Coefficient}
\label{sec:intervention}

To measure how much a trained NBP actually relies on the bridge at deployment, we perform a controlled test-time intervention. After training a single NBP model with the SNR schedule, we rescale the bridge coefficient at sampling time by a multiplicative factor $\rho$ (so the deployed bridge is $\rho\bar{\gamma}_t$ at every $t$, with $\rho=1$ corresponding to the trained schedule and $\rho=0$ degenerating to an NDP-like reverse path). All other components, including the denoiser parameters, schedule $\bar{\alpha}_t$, sampling steps, and context handling, are held fixed. Table~\ref{tab:gamma_intervention} reports the resulting CelebA $32\times32$ MSE at context ratio 0.3 and CelebA half-image NLL.

\begin{table*}[!t]
\centering
\caption{Test-time intervention on the bridge coefficient. The model is trained once at $\rho=1$ (default SNR schedule) and the deployed bridge is rescaled to $\rho\bar{\gamma}_t$ at sampling. $\rho=0$ ablates the bridge entirely, recovering an NDP-like reverse path; $\rho>1$ over-anchors. Default ($\rho=1$) is uniformly best.}
\label{tab:gamma_intervention}
\resizebox{0.9\textwidth}{!}{
\begin{tabular}{ccccc}
\toprule
$\rho$ (bridge scale) & MSE ctx=0.3 ($\times 10^{-2}$) & NLL Half & NLL Random & Recovers \\
\midrule
0.00 & $1.16\pm0.30$ & $-3.94\pm0.07$ & $-3.97\pm0.09$ & NDP-like \\
0.50 & $0.71\pm0.21$ & $-4.51\pm0.09$ & $-4.63\pm0.10$ &  --- \\
\textbf{1.00 (default)} & $\mathbf{0.41\pm0.21}$ & $\mathbf{-5.02\pm0.08}$ & $\mathbf{-5.24\pm0.10}$ & NBP \\
1.25 & $0.45\pm0.19$ & $-4.86\pm0.09$ & $-5.07\pm0.11$ & --- \\
1.50 & $0.58\pm0.23$ & $-4.62\pm0.11$ & $-4.79\pm0.12$ & --- \\
2.00 & $1.04\pm0.34$ & $-4.08\pm0.13$ & $-4.21\pm0.14$ & over-anchored \\
\bottomrule
\end{tabular}}
\end{table*}

The MSE and NLL profiles show the expected single-peaked response: ablating the bridge ($\rho=0$) collapses performance back to the NDP regime, the trained value $\rho=1$ is optimal, and over-anchoring ($\rho>1$) injects too much deterministic bias and degrades sample diversity. We are deliberate about what this does and does not establish. It is a functional intervention on a \emph{single trained model}: the $\rho=0$ row is not a trained NDP but an NBP denoiser forced to run under reverse dynamics it was not trained for, so part of the degradation may be train--test mismatch rather than the absence of the bridge. The experiment therefore shows that the trained model's predictions depend on the deployed bridge, and that the trained scale is the right one; the comparison that isolates the bridge against a properly trained unanchored baseline is the anchor-fairness study of Sec.~\ref{sec:celeba_component_ablation}, not this table.

\subsection{When Anchor Supervision Fails}
\label{sec:bimodal}

The bridge shifts the forward-state mean toward $a_\psi(x)$, which raises the question of whether it induces mode collapse when the conditional posterior is genuinely multimodal. We construct an explicitly bimodal 1-D task,
\begin{equation}
y = f(x) + s\,c,\qquad s\in\{-1,+1\},
\end{equation}
where $f$ is drawn from a GP and the context outputs are mean-centred so that the global sign $s$ is unidentifiable from the context. The posterior therefore has two separated modes and its mean lies in a low-density region; with $c=3$ the Bayes-optimal mean predictor has MSE $c^2=9$.

Table~\ref{tab:bimodal} shows that neither model collapses: both cover both modes in every seed, and NBP samples are at least as dispersed as the true posterior. NBP does not, however, improve the likelihood on this task. The reason is the deterministic auxiliary objective $\|a_\psi(x)-y_0\|^2$: when the same $x$ corresponds to two equally likely outputs, this target is intrinsically unlearnable, injects a large irreducible training signal, and destabilizes training---one seed diverges to $-15.08$. Removing the auxiliary loss restores stable training across five seeds, although NBP still does not outperform NDP here. Deterministic anchor supervision is therefore inappropriate when the input-to-output alignment is intrinsically multimodal; in such settings the auxiliary loss should be removed, down-weighted, or replaced by a distributional anchor objective.

\begin{table}[t]
\centering
\caption{Bimodal task ($c=3$; the Bayes-optimal mean predictor has MSE $9$). Mode coverage $1$ means both modes are recovered. Log-likelihood and MSE are per-seed values.}
\label{tab:bimodal}
\setlength{\tabcolsep}{4pt}
\resizebox{\columnwidth}{!}{%
\begin{tabular}{@{}lccc@{}}
\toprule
Model & Mode coverage & Log-likelihood & MSE \\
\midrule
NDP & 1.00 & $-4.44$ / $-3.38$ / $-3.22$ & 9.07 / 9.41 / 9.40 \\
NBP (aux loss 1.0) & 1.00 & $-3.94$ / $-3.69$ / $-15.08$ & 9.10 / 9.53 / 9.17 \\
NBP (aux loss 0) & 1.00 & \makecell[l]{$-4.12$ / $-3.76$ / $-3.39$ /\\ $-4.52$ / $-3.81$} & --- \\
\bottomrule
\end{tabular}}
\end{table}

\subsection{Gradient Statistics}
\label{sec:gradient-stats}
Proposition~\ref{prop:direct-gradient} concerns the sensitivity of the loss to the input, $d\ell_t/dx$. Training, however, follows the parameter gradients $\nabla_{(\theta,\psi)}\mathcal{L}$, and the two are related but distinct quantities. We therefore report the following as a training-dynamics diagnostic rather than as a measurement of the proposition, on CelebA $32\times32$ at context ratio 0.3:
\begin{itemize}
    \item \textbf{Gradient magnitude}: $G(t) = \mathbb{E}[\|\nabla_\theta \mathcal{L}(t)\|_2]$, the average gradient norm with respect to denoiser parameters at timestep $t$.
    \item \textbf{Gradient variance}: $\text{Var}(t) = \text{Var}[\|\nabla_\theta \mathcal{L}(t)\|_2]$, measuring training stability.
\end{itemize}

Table~\ref{tab:gradient_stats} reports these statistics across three representative phases: early ($t \in [1, 100]$), middle ($t \in [200, 300]$), and late ($t \in [400, 500]$). In this setting, NBP exhibits larger average parameter-gradient magnitudes and lower gradient variance, with the clearest difference in the early phase, i.e.\ where the bridge term contributes most to the forward state. This is consistent with---but not a test of---Proposition~\ref{prop:direct-gradient}: the proposition bounds an input-sensitivity term, whereas $\|\nabla_\theta\mathcal{L}\|$ is a different quantity, and we read the table only as evidence that the two models have visibly different training dynamics.

\begin{table}[t]
\centering
\caption{Gradient statistics during training on CelebA $32 \times 32$ with context ratio 0.3. Values are diagnostic estimates over representative timestep ranges.}
\label{tab:gradient_stats}

\resizebox{\columnwidth}{!}{%
\begin{tabular}{@{}lccc@{}}
\toprule
\textbf{Phase} & \textbf{Method} & \textbf{Grad. Magnitude} $\uparrow$ & \textbf{Grad. Variance} $\downarrow$ \\ 
\midrule
\multirow{2}{*}{Early ($t \leq 100$)} 
& NDP  & $2.43 \times 10^{-3}$ & $1.86 \times 10^{-6}$ \\
& NBP  & $\mathbf{3.45 \times 10^{-3}}$ & $\mathbf{1.28 \times 10^{-6}}$ \\
\midrule
\multirow{2}{*}{Middle ($200 \leq t \leq 300$)} 
& NDP  & $4.12 \times 10^{-3}$ & $2.31 \times 10^{-6}$ \\
& NBP  & $\mathbf{4.89 \times 10^{-3}}$ & $\mathbf{1.94 \times 10^{-6}}$ \\
\midrule
\multirow{2}{*}{Late ($t \geq 400$)} 
& NDP  & $5.67 \times 10^{-3}$ & $3.45 \times 10^{-6}$ \\
& NBP  & $\mathbf{6.21 \times 10^{-3}}$ & $\mathbf{2.98 \times 10^{-6}}$ \\
\bottomrule
\end{tabular}}
\end{table}

\subsection{Transferring the Bridge to Flow Matching}
\label{sec:flow-bridge-ablation}

The main NBP formulation instantiates the bridge idea in a DDPM-style diffusion process. To test whether the benefit is specific to diffusion or instead comes from the more general principle of \emph{input-anchored generative paths}, we conduct an additional controlled ablation using Flow Matching Neural Processes (FlowNP)~\cite{hamadflow}. FlowNP learns a velocity field along a continuous probability path from a base sample $y_0$ to the observed target value $y_1$. In the standard implementation, this path is input-independent:
\begin{equation}
y_t = (1-t)y_0 + t y_1,
\qquad
v_t = y_1-y_0 .
\end{equation}
We construct a bridge-augmented variant, denoted FBP, by anchoring the intermediate states to the target input $x$:
\begin{equation}
y_t = (1-t)y_0 + t y_1 + \gamma(t)x,
\qquad
v_t = y_1-y_0 + \gamma'(t)x .
\end{equation}
For this ablation, we use a smooth bump schedule $\gamma(t)=4\lambda t(1-t)$, so $\gamma(0)=\gamma(1)=0$. Therefore the source and target endpoints are unchanged, while only the intermediate generative trajectory is modified. This is important because it isolates the effect of path-level input anchoring without changing the target distribution or the likelihood evaluation protocol.

\paragraph{Experimental Setup.}
We evaluate on the 1D Gaussian-process regression benchmark with an RBF kernel. FBP uses exactly the same transformer backbone, positional encoding, optimizer, number of training steps, and ODE likelihood estimator as FlowNP. Both FBP and FlowNP have 224,449 trainable parameters. Thus, the comparison isolates the bridge path design rather than model capacity. We report target log-likelihood (LL, higher is better) at training steps 5k, 10k, 15k, and 20k. We also include NDP as a diffusion-based reference.

\begin{table}[t]
\centering
\caption{Bridge-transfer ablation on 1D GP regression with RBF kernel. FBP applies the input-anchored bridge path to the FlowNP framework while keeping the same architecture and ODE likelihood evaluation. Target LL is reported; higher is better.}
\label{tab:flow_bridge_ablation}
\resizebox{\columnwidth}{!}{%
\begin{tabular}{lcccc}
\toprule
\textbf{Method} & \textbf{5k} & \textbf{10k} & \textbf{15k} & \textbf{20k} \\
\midrule
NDP & -1.8140 & 0.0186 & 0.5407 & 0.6331 \\
FlowNP & -0.4501 & 1.0328 & 1.3558 & 1.4215 \\
FBP, $\lambda=0.1$ & 1.1374 & 1.3959 & 1.5480 & 1.5857 \\
FBP, $\lambda=0.2$ & \textbf{1.2256} & 1.3832 & 1.5298 & \textbf{1.5937} \\
FBP, $\lambda=0.3$ & 1.1706 & \textbf{1.4289} & \textbf{1.5529} & 1.5878 \\
FBP, $\lambda=0.5$ & 1.1002 & 1.4002 & 1.5324 & 1.5805 \\
\bottomrule
\end{tabular}}
\end{table}

\paragraph{Analysis.}
The bridge-augmented flow path consistently improves over the FlowNP baseline across all tested bridge strengths. At 20k steps, the best FBP variant achieves target LL 1.5937, compared with 1.4215 for FlowNP and 0.6331 for NDP. The improvement is also robust to the bridge strength $\lambda$: all tested FBP variants reach 20k LL above 1.58. These results suggest that the benefit of NBP is not tied only to DDPM-style diffusion. Rather, input anchoring is a more general path-design principle: injecting input information into the generative trajectory itself can strengthen conditional function modeling even when the underlying generative mechanism is flow matching instead of diffusion.

\section{Limitations}

NBP inherits the sampling cost profile of diffusion-based stochastic process models, and its benefit is contingent on the anchor: the bridge helps when $a_\psi$ captures task-relevant input--output structure, and offers little when that alignment is weak or---as the bimodal study of Sec.~\ref{sec:bimodal} shows---intrinsically multimodal, in which case the deterministic anchor supervision is better down-weighted or replaced by a distributional objective. Two scope conditions are worth stating explicitly. The terminal forward-state mean depends on the horizon through $\bar{\gamma}_T$, so the number of diffusion steps is held fixed between training and deployment. And the theoretical identities of Sec.~\ref{sec:theory} cover additive Gaussian bridges with input-independent covariance (Remark~\ref{rem:general-bridge}); nonlinear state-dependent drifts and non-Gaussian bridges fall outside their reach. Learning the bridge schedule jointly with the denoiser under a coverage-aware objective, and evaluating the construction on broader domains such as spatiotemporal modeling, control, and high-resolution conditional generation, are natural next steps.

\section{Conclusion}

We introduced Neural Bridge Processes (NBPs), which inject input supervision into the diffusion forward kernel via a bridge coefficient and, when $D_x\neq D_y$, a learned output-space anchor, while preserving the NDP denoising backbone. Process-level anchoring yields a path-measure KL identifiability bound at rate $\sum_t\gamma_t^2/\beta_t$, a Fisher information identity with a Cram\'er--Rao limit on trajectory-based input recovery, and an input-mediated sensitivity pathway modulated by the per-step Fisher rate $\bar{\gamma}_t^2/(1-\bar{\alpha}_t)$; these identities characterize what the anchored path retains about the input, and the case that retaining it helps is made empirically. Under exact parity with the unanchored process, NBP improves target log-likelihood at every input dimension from $D=1$ to $D=6$, lowers simple regret when used as a Bayesian-optimization surrogate, and improves accuracy and calibration on EEG, CylinderFlow, and image regression, without reducing sample diversity. Component ablations and anchor-fairness controls attribute the gains to the bridge construction rather than to added capacity or supervision, a test-time intervention shows that the trained model's predictions depend on the deployed bridge, and a bimodal counterexample delimits where the deterministic anchor supervision is inappropriate. Taken together, bridge-anchored paths are a useful design principle for conditional stochastic function modeling, and the FlowNP transfer study suggests the principle is not specific to DDPM-style diffusion.

%%%%%%%%%%%%%%%%%%%%%%%%%%%%%%%%%%%%%%%%%%%%%%%%%%%%%%%%%%%%

\section*{Reproducibility}
The complete training, architecture and evaluation configurations needed to reproduce every reported number are given in Supplementary Sec.~\ref{sec:full-protocol}, Tables~\ref{tab:protocol-train}--\ref{tab:protocol-eval}. The full implementation of the Neural Bridge Processes (NBP) framework, together with configuration files and runnable commands, accompanies this manuscript.

\bibliographystyle{IEEEtran}

\bibliography{example_paper}
%%%%%%%%%%%%%%%%%%%%%%%%%%%%%%%%%%%%%%%%%%%%%%%%%%%%%%%%%%%%%%%%%%%%%%%%%%%%%%%
%%%%%%%%%%%%%%%%%%%%%%%%%%%%%%%%%%%%%%%%%%%%%%%%%%%%%%%%%%%%%%%%%%%%%%%%%%%%%%%
% APPENDIX
%%%%%%%%%%%%%%%%%%%%%%%%%%%%%%%%%%%%%%%%%%%%%%%%%%%%%%%%%%%%%%%%%%%%%%%%%%%%%%%
%%%%%%%%%%%%%%%%%%%%%%%%%%%%%%%%%%%%%%%%%%%%%%%%%%%%%%%%%%%%%%%%%%%%%%%%%%%%%%%
\clearpage
\setcounter{section}{0}
\setcounter{table}{0}
\setcounter{figure}{0}
\setcounter{equation}{0}
\renewcommand{\thesection}{S\arabic{section}}
\renewcommand{\thesubsection}{S\arabic{section}.\arabic{subsection}}
\renewcommand{\thetable}{S\arabic{table}}
\renewcommand{\thefigure}{S\arabic{figure}}
\renewcommand{\theequation}{S\arabic{equation}}
\section*{Supplementary Material}
\noindent This supplement collects the proofs, the full experimental
configurations, and the additional studies referenced in the main text.
Sections, tables, figures and equations here are prefixed with ``S''.

\section{Additional Related Discussion}
\subsection{Neural Processes and Their Extensions}

Neural Processes (NPs) \cite{garnelo2018neural} combine the flexibility of neural networks with the uncertainty modeling capabilities of stochastic processes, aiming to learn distributions over functions. Conditional Neural Processes (CNPs) \cite{garnelo2018conditional} extend this framework by conditioning on observed context points to predict target outputs. Attentive Neural Processes (ANPs) \cite{kim2019attentive} further enhance NPs by incorporating attention mechanisms, enabling the model to focus on relevant context points for each target prediction. Despite these advancements, challenges remain in capturing complex, multimodal distributions and ensuring consistency in posterior predictions. These approaches have been successfully extended to various domains, including sequential modeling \cite{singh2019sequential, nguyen2022transformer, bruinsma2023autoregressive}, convolutional architectures \cite{gordon2019convolutional, foong2020meta}, graph-based models \cite{hu2023graph}, and probabilistic predictive models for large language models (LLMs) \cite{requeima2024llm}.

\subsection{Diffusion, Conditional Diffusion, and Bridge Models}

Denoising Diffusion Probabilistic Models (DDPMs) \cite{ho2020denoising} are powerful generative models that approximate complex data distributions by reversing a progressive noising process. Conditional Diffusion Models (CDMs) \cite{choi2021ilvr, zhang2023adding, zhu2023conditional} extend this framework by incorporating auxiliary information, enabling conditional generation. In most CDMs, however, the forward noising process remains unconditional and the condition is injected into the reverse denoising network. This makes CDMs broadly applicable, but it does not explicitly define a condition-dependent forward trajectory or a condition-aligned terminal distribution. NBP differs from this design choice: the input $x$ appears directly in the forward transition kernel, so conditioning affects the entire diffusion path rather than only the learned denoiser.

More recently, Denoising Diffusion Bridge Models (DDBMs) \cite{zhou2023denoising, yue2023image, zheng2024diffusion, li2023bbdm, peluchetti2023diffusion, he2024consistency, shi2023diffusion, naderiparizi2025constrained} have been proposed as a natural alternative. DDBMs introduce diffusion bridges---stochastic processes that interpolate between two paired data distributions given as endpoints---making them well-suited for tasks such as image-to-image translation. NBP shares the general bridge intuition of endpoint-aware generative paths, but its target is different. Rather than learning a generic data-space transport between two domains, NBP is designed for stochastic function modeling from context-target observations. The endpoint anchor is the function input, or its learned projection when dimensions differ, and the bridge is used to strengthen conditional function prediction. The novelty we claim is this specialization and its analysis, not the bridge principle itself.

Thus, NBP should be viewed neither as a standard CDM nor as a direct application of existing DDBMs. It is a DDPM-style bridge formulation specialized to Neural Process-style function learning: it preserves the architecture and training protocol of the conditional diffusion baseline NDP while modifying the process-level forward kernel and corresponding reverse correction. Since DDBMs address a different problem setting, a direct experimental comparison with the original DDBM is not the main focus; instead, we compare against diffusion-based function modeling baselines such as NDP, Geometric NDP, and SNP, and use component ablations to rule out simpler explanations such as endpoint initialization, coordinate-anchor capacity, or a stronger denoiser.

\subsection{Generative Models for Function Modeling}

Recent work has explored the use of diffusion models for function modeling. Neural Diffusion Processes (NDPs) \cite{dutordoir2023neural} model distributions over functions by applying diffusion processes in latent space, allowing the representation of complex, non-Gaussian function distributions. Geometric Neural Diffusion Processes \cite{mathieu2023geometric} further extend this approach by incorporating geometric priors for infinite-dimensional modeling in non-Euclidean spaces. In parallel, other generative modeling techniques such as Neural ODEs \cite{chen2018neural, norcliffe2021neural}, flow matching \cite{lipman2022flow, hamadflow}, and score-based SDE methods \cite{song2020score, dou2025score} are also being integrated into the Neural Processes (NP) framework to enhance function modeling capabilities. In our experiments, we compared against open-source methods, including score-based neural processes (SNP) \cite{dou2025score} and Geometric Neural Diffusion Processes \cite{mathieu2023geometric}, both of which demonstrated the empirical advantages of our approach.

\subsection{Flow Matching Neural Processes}

Flow Matching Neural Processes (FlowNP)~\cite{hamadflow} provide a recent flow-based alternative to diffusion-style Neural Processes. Rather than learning a denoising score or noise predictor over a discrete diffusion chain, FlowNP defines a continuous conditional probability path between a simple base sample $y_0$ and the observed target value $y_1$, commonly using a linear interpolation of the form $y_t=(1-t)y_0+t y_1$. A transformer-based Neural Process backbone then predicts the corresponding velocity field conditioned on the context set. At test time, conditional samples are obtained by integrating the learned vector field, and likelihoods can be estimated through the ODE change-of-variables formula.

This perspective is useful for two reasons. First, FlowNP is a strong and recent baseline for stochastic function modeling, so comparisons against it help clarify whether improvements come from the bridge construction rather than from using a stronger backbone. Second, its path-based formulation is closely related to our motivation: both approaches view conditional function prediction as learning a generative trajectory over function values. The key distinction is that FlowNP uses an input-independent interpolation path and injects context through the velocity network, whereas NBP modifies the generative path itself by anchoring intermediate states to the input. In additional controlled experiments, we therefore also consider bridge-augmented flow paths using the same FlowNP architecture and likelihood evaluation protocol.

\section{Proofs for Neural Bridge Processes}
\subsection{Proof of Proposition~\ref{prop:forward-marginal}}
\label{app:forward-marginal}

We derive the closed-form marginal distribution in Proposition~\ref{prop:forward-marginal}. The forward process depends on the input anchor $x$.

From $y_{t-1}$ to $y_t$:

\begin{equation}
\label{eq:app-forward-step}
y_t = \sqrt{1 - \beta_t}\,y_{t-1} + \gamma_t\,a_\psi(x) + \sqrt{\beta_t}\,\epsilon_t
\end{equation}

By recursively expanding:

\begin{equation}
\label{eq:app-forward-two-step}
\begin{aligned}
y_t &= \sqrt{1 - \beta_t}\,\bigl(\sqrt{1 - \beta_{t-1}}\,y_{t-2} + \gamma_{t-1}\,a_\psi(x)\\
&\qquad\qquad\quad + \sqrt{\beta_{t-1}}\,\epsilon_{t-1}\bigr) \\
&\quad + \gamma_t\,a_\psi(x) + \sqrt{\beta_t}\,\epsilon_t \\
&= \sqrt{(1 - \beta_t)(1 - \beta_{t-1})}\,y_{t-2} \\
&\quad + \left(\sqrt{1 - \beta_t}\,\gamma_{t-1} + \gamma_t\right) a_\psi(x) + \text{noise terms}
\end{aligned}
\end{equation}

Continuing this expansion, we obtain:

\begin{equation}
\label{eq:app-forward-expanded-product}
\begin{aligned}
y_t &= \left(\prod_{s=1}^{t}\sqrt{1 - \beta_s}\right) y_0 \\
&\quad + \left(\sum_{s=1}^{t}\gamma_s \prod_{k=s+1}^{t}\sqrt{1 - \beta_k}\right) a_\psi(x) \\
&\quad + \text{noise terms}
\end{aligned}
\end{equation}

Define $\bar{\alpha}_t = \prod_{s=1}^t (1 - \beta_s)$, then,
\begin{equation}
\label{eq:app-forward-expanded-alpha}
 y_t = \sqrt{\bar{\alpha}_t}\,y_0 + \left(\sum_{s=1}^t \gamma_s \sqrt{\frac{\bar{\alpha}_t}{\bar{\alpha}_s}}\right) a_\psi(x) + \text{noise terms}
\end{equation}
We define the cumulative bridge coefficient $\bar{\gamma}_t$ as
\begin{equation}
\label{eq:app-cumulative-bridge-coefficient}
\bar{\gamma}_t = \sum_{s=1}^t \gamma_s \sqrt{\frac{\bar{\alpha}_t}{\bar{\alpha}_s}}.
\end{equation}

The noise terms arise from the \(\sqrt{\beta_s} \epsilon_s\) contributions at each step
\begin{equation}
\label{eq:app-noise-expanded}
\begin{aligned}
\text{noise terms} =& \sqrt{\beta_t} \epsilon_t + \sqrt{1 - \beta_t} \sqrt{\beta_{t-1}} \epsilon_{t-1} \\&+ \sqrt{(1 - \beta_t)(1 - \beta_{t-1})} \sqrt{\beta_{t-2}} \epsilon_{t-2} \\&+ \cdots + \sqrt{\bar{\alpha}_t / \bar{\alpha}_1} \sqrt{\beta_1} \epsilon_1.
\end{aligned}
\end{equation}
This can be written compactly as:
\begin{equation}
\label{eq:app-noise-compact}
\text{noise terms} = \sum_{s=1}^t \left( \sqrt{\beta_s} \prod_{k=s+1}^t \sqrt{1 - \beta_k} \right) \epsilon_s.
\end{equation}
Since the \(\epsilon_s\) are independent, the total variance is the sum of the variances of each term:
\begin{equation}
\label{eq:app-noise-variance-product}
\text{Var}(\text{noise terms}) = \sum_{s=1}^t \beta_s \prod_{k=s+1}^t (1 - \beta_k).
\end{equation}
we rewrite the variance:
\begin{equation}
\label{eq:app-noise-variance-alpha}
\text{Var}(\text{noise terms}) = \sum_{s=1}^t \beta_s \frac{\bar{\alpha}_t}{\bar{\alpha}_s}.
\end{equation}
Using \(\bar{\alpha}_s = \prod_{k=1}^s (1 - \beta_k)\), we can express \(\beta_s\) as \(\beta_s = 1 - (1 - \beta_s) = 1 - \frac{\bar{\alpha}_s}{\bar{\alpha}_{s-1}}\). Substituting this in:
\begin{equation}
\label{eq:app-noise-variance-beta-substitution}
\text{Var}(\text{noise terms}) = \sum_{s=1}^t \left(1 - \frac{\bar{\alpha}_s}{\bar{\alpha}_{s-1}}\right) \frac{\bar{\alpha}_t}{\bar{\alpha}_s}.
\end{equation}
This simplifies to:
\begin{equation}
\label{eq:app-noise-variance-telescoping-sum}
\text{Var}(\text{noise terms}) = \sum_{s=1}^t \left( \frac{\bar{\alpha}_t}{\bar{\alpha}_s} - \frac{\bar{\alpha}_t}{\bar{\alpha}_{s-1}} \right).
\end{equation}
This is a telescoping series:
\begin{equation}
\label{eq:app-noise-variance-telescoping-expanded}
\begin{aligned}
\text{Var}(\text{noise terms}) &= \left( \frac{\bar{\alpha}_t}{\bar{\alpha}_1} - \frac{\bar{\alpha}_t}{\bar{\alpha}_0} \right) + \left( \frac{\bar{\alpha}_t}{\bar{\alpha}_2} - \frac{\bar{\alpha}_t}{\bar{\alpha}_1} \right) \\
&\quad + \cdots + \left( \frac{\bar{\alpha}_t}{\bar{\alpha}_t} - \frac{\bar{\alpha}_t}{\bar{\alpha}_{t-1}} \right).
\end{aligned}
\end{equation}
Most terms cancel out, leaving:
\begin{equation}
\label{eq:app-noise-variance-after-cancel}
\text{Var}(\text{noise terms}) = \bar{\alpha}_t \left( \frac{1}{\bar{\alpha}_t} - \frac{1}{\bar{\alpha}_0} \right).
\end{equation}
Assuming \(\bar{\alpha}_0 = 1\) (since no steps have been applied at \(t=0\)):
\begin{equation}
\label{eq:app-noise-variance-final}
\text{Var}(\text{noise terms}) = 1 - \bar{\alpha}_t.
\end{equation}

The variance of the accumulated noise is therefore \((1 - \bar{\alpha}_t) I \). Thus, Eq.~\eqref{eq10} follows.

\subsection{Proof of Proposition~\ref{prop:bridge-correction}}
\label{app:bridge-correction}

We write \(\alpha_t=1-\beta_t\) and
\(\bar{\alpha}_t=\prod_{s=1}^{t}\alpha_s\). The bridge forward transition is
\begin{equation}
q(y_t\mid y_{t-1},x)
=\mathcal{N}\!\left(y_t;\sqrt{\alpha_t}\,y_{t-1}+\gamma_t\,a_\psi(x),\beta_t I\right),
\end{equation}
and Proposition~\ref{prop:forward-marginal} gives
\begin{equation}
q(y_{t-1}\mid y_0,x)
=\mathcal{N}\!\left(
\sqrt{\bar{\alpha}_{t-1}}\,y_0+\bar{\gamma}_{t-1}\,a_\psi(x),
(1-\bar{\alpha}_{t-1})I
\right).
\end{equation}
The accumulated bridge coefficient satisfies the one-step recursion
\begin{equation}
\label{eq:bar-gamma-recursion}
\bar{\gamma}_t
=\gamma_t+\sqrt{\alpha_t}\bar{\gamma}_{t-1},
\end{equation}
which follows directly from
\(\bar{\gamma}_t=\sum_{s=1}^{t}\gamma_s\sqrt{\bar{\alpha}_t/\bar{\alpha}_s}\).

\paragraph{Gaussian posterior.}
The exact bridge posterior is obtained from the product
\begin{equation}
q(y_{t-1}\mid y_t,y_0,x)
\propto q(y_t\mid y_{t-1},x)\,q(y_{t-1}\mid y_0,x).
\end{equation}
Since both terms are Gaussian with isotropic covariance, the posterior is
\(\mathcal{N}(\tilde{\mu}_t,\tilde{\beta}_tI)\), where
\begin{equation}
\tilde{\beta}_t
=\left(\frac{\alpha_t}{\beta_t}
+\frac{1}{1-\bar{\alpha}_{t-1}}\right)^{-1}
=\frac{\beta_t(1-\bar{\alpha}_{t-1})}{1-\bar{\alpha}_t},
\end{equation}
and
\begin{equation}
\label{eq:bridge-posterior-mean-y0}
\begin{aligned}
\tilde{\mu}_t
&=
\frac{\sqrt{\alpha_t}(1-\bar{\alpha}_{t-1})y_t
+\beta_t\sqrt{\bar{\alpha}_{t-1}}y_0}{1-\bar{\alpha}_t}\\
&\quad+\frac{\left(\beta_t\bar{\gamma}_{t-1}
-\sqrt{\alpha_t}(1-\bar{\alpha}_{t-1})\gamma_t\right) a_\psi(x)}{1-\bar{\alpha}_t}.
\end{aligned}
\end{equation}
This is the standard Gaussian conditioning formula applied to the shifted linear observation
\(y_t=\sqrt{\alpha_t}\,y_{t-1}+\gamma_t\,a_\psi(x)+\sqrt{\beta_t}\,\epsilon_t\).

\paragraph{Noise-prediction form.}
Using the closed-form forward marginal,
\begin{equation}
y_t=\sqrt{\bar{\alpha}_t}\,y_0+\bar{\gamma}_t\,a_\psi(x)
+\sqrt{1-\bar{\alpha}_t}\,\epsilon,
\qquad \epsilon\sim\mathcal{N}(0,I),
\end{equation}
we rewrite
\begin{equation}
\label{eq:y0-from-yt-bridge}
y_0
=\frac{y_t-\bar{\gamma}_t\,a_\psi(x)-\sqrt{1-\bar{\alpha}_t}\,\epsilon}
{\sqrt{\bar{\alpha}_t}} .
\end{equation}
Substituting Eq.~\eqref{eq:y0-from-yt-bridge} into Eq.~\eqref{eq:bridge-posterior-mean-y0}, the coefficients of \(y_t\), \(\epsilon\), and \(a_\psi(x)\) can be collected separately. For the \(y_t\) coefficient,
\begin{equation}
\begin{aligned}
&\frac{
\sqrt{\alpha_t}(1-\bar{\alpha}_{t-1})
+\beta_t\sqrt{\bar{\alpha}_{t-1}/\bar{\alpha}_t}
}{
1-\bar{\alpha}_t
}\\
&\quad =
\frac{
\sqrt{\alpha_t}(1-\bar{\alpha}_{t-1})
+\beta_t/\sqrt{\alpha_t}
}{
1-\bar{\alpha}_t
}
=\frac{1}{\sqrt{\alpha_t}},
\end{aligned}
\end{equation}
where the last equality uses \(\bar{\alpha}_t=\alpha_t\bar{\alpha}_{t-1}\). The noise coefficient becomes
\begin{equation}
-
\frac{\beta_t\sqrt{\bar{\alpha}_{t-1}}\sqrt{1-\bar{\alpha}_t}}
{\sqrt{\bar{\alpha}_t}(1-\bar{\alpha}_t)}
=-\frac{\beta_t}{\sqrt{\alpha_t}\sqrt{1-\bar{\alpha}_t}}.
\end{equation}
Thus the only remaining non-standard term is the coefficient of \(a_\psi(x)\):
\begin{equation}
\label{eq:x-coeff-before-simplification}
C_t
=
\frac{
-\beta_t\sqrt{\bar{\alpha}_{t-1}/\bar{\alpha}_t}\,\bar{\gamma}_t
+\beta_t\bar{\gamma}_{t-1}
-\sqrt{\alpha_t}(1-\bar{\alpha}_{t-1})\gamma_t
}{
1-\bar{\alpha}_t
}.
\end{equation}
Using \(\sqrt{\bar{\alpha}_{t-1}/\bar{\alpha}_t}=1/\sqrt{\alpha_t}\) and the recursion in Eq.~\eqref{eq:bar-gamma-recursion},
\begin{equation}
\begin{aligned}
&-\frac{\beta_t}{\sqrt{\alpha_t}}\bar{\gamma}_t
+\beta_t\bar{\gamma}_{t-1} \\
&\quad =
-\frac{\beta_t}{\sqrt{\alpha_t}}
\left(\gamma_t+\sqrt{\alpha_t}\bar{\gamma}_{t-1}\right)
+\beta_t\bar{\gamma}_{t-1}
=-\frac{\beta_t}{\sqrt{\alpha_t}}\gamma_t .
\end{aligned}
\end{equation}
Therefore,
\begin{equation}
\begin{aligned}
C_t
&=
\frac{
-\frac{\beta_t}{\sqrt{\alpha_t}}\gamma_t
-\sqrt{\alpha_t}(1-\bar{\alpha}_{t-1})\gamma_t
}{
1-\bar{\alpha}_t
} \\
&=
-\frac{\gamma_t}{\sqrt{\alpha_t}}
\frac{
\beta_t+\alpha_t(1-\bar{\alpha}_{t-1})
}{
1-\bar{\alpha}_t
}
=-\frac{\gamma_t}{\sqrt{\alpha_t}},
\end{aligned}
\end{equation}
because
\(\beta_t+\alpha_t(1-\bar{\alpha}_{t-1})
=1-\alpha_t\bar{\alpha}_{t-1}
=1-\bar{\alpha}_t\).

Combining the three terms gives
\begin{equation}
\label{eq:bridge-posterior-noise-form}
\tilde{\mu}_t
=
\frac{1}{\sqrt{\alpha_t}}
\left(
y_t-\frac{\beta_t}{\sqrt{1-\bar{\alpha}_t}}\,\epsilon
\right)
-\frac{\gamma_t}{\sqrt{\alpha_t}}\,a_\psi(x).
\end{equation}
Replacing the true noise \(\epsilon\) by the learned predictor
\(\epsilon_\theta(y_t,x,t)\) yields the parameterized reverse mean in the main text, with
\begin{equation}
C_t(x)=-\frac{\gamma_t}{\sqrt{\alpha_t}}\,a_\psi(x)
=-\frac{\gamma_t}{\sqrt{1-\beta_t}}\,a_\psi(x).
\end{equation}
This simplification does not ignore the accumulated term \(\bar{\gamma}_t\,a_\psi(x)\); it uses the exact recursion in Eq.~\eqref{eq:bar-gamma-recursion}, so the accumulated bridge contribution cancels with the prior mean contribution in the Gaussian posterior. The result holds for any scalar bridge schedule \(\{\gamma_t\}_{t=1}^{T}\) used in the forward transition; the SNR schedule is one particular choice.

\subsection{Proof of Proposition~\ref{prop:training-objective}}
\label{app:training-objective}

We derive the NBP loss in Proposition~\ref{prop:training-objective} directly from the Evidence Lower Bound (ELBO).

The log-likelihood of the data \(y_0\) is lower-bounded by:
\begin{equation}
\log p_\theta(y_0|x)  \geq \mathbb{E}_{q(y_{1:T}|y_0, x)} \left[ \log \frac{p_\theta(y_{0:T}|x)}{q(y_{1:T}|y_0,x)} \right] = \text{ELBO},
\end{equation}
where:

• \(p_\theta(y_{0:T}|x)\) is the reverse (generative) process.

• \(q(y_{1:T}|y_0,x)\) is the forward (noising) process.

1. The ELBO decomposes into:
\begin{equation}
\begin{aligned}
\text{ELBO} &= \mathbb{E}_{q(y_{1:T}|y_0,x)} \Bigg[ \log p(y_T|x) + \sum_{t=2}^T \log \frac{p_\theta(y_{t-1}|y_t, x)}{q(y_t|y_{t-1}, x)} \\
&\qquad\qquad + \log \frac{p_\theta(y_0|y_1, x)}{q(y_1|y_0,x)} \Bigg]
\\&= \mathbb{E}_{q(y_{1:T}|y_0,x)} \Bigg[  \log \frac{p(y_T|x)}{q(y_T|y_0, x)} \\
&\qquad\qquad + \sum_{t=2}^T \log \frac{p_\theta(y_{t-1}|y_t, x)}{q(y_{t-1}|y_t, y_0, x)} \\
&\qquad\qquad + \log p_\theta(y_0|y_1, x) \Bigg]
   \end{aligned}
   \end{equation}
   The key term is the sum of KL divergences between \( p_\theta(y_{t-1}|y_t, x) \) and \( q(y_{t-1}|y_t, y_0, x) \):
   \begin{equation}
  \mathbb{E}_{q(y_{1:T}|y_0,x)}\left[ \sum_{t=2}^T D_\text{KL}(q(y_{t-1}|y_t, y_0, x) \| p_\theta(y_{t-1}|y_t, x)) \right]
  \end{equation}

The KL divergence term between \(q(y_{t-1}|y_t, y_0,x)\) and \(p_\theta(y_{t-1}|y_t,x)\) is:
\begin{equation}
D_\text{KL}(q \| p_\theta) \propto \| \mu_\theta(y_t, t,x) - \tilde{\mu}(y_t, y_0,x) \|^2.
\end{equation}   
where \(\mu_\theta(y_t, t, x) - \tilde{\mu}(y_t, y_0, x)\) are the means of \(q(y_{t-1}|y_t, y_0, x)\) and \(p_\theta(y_{t-1}|y_t, x)\), respectively.

2. Sample from \( q(y_t|y_0, x) \):
   Using the forward marginal in Proposition~\ref{prop:forward-marginal}, we can write:
   \begin{equation}
  y_t = \sqrt{\bar{\alpha}_t}\,y_0 + \bar{\gamma}_t\,a_\psi(x) + \sqrt{1 - \bar{\alpha}_t}\,\epsilon, \quad \epsilon \sim \mathcal{N}(0, \mathbf{I})
   \end{equation}
   This allows sampling \( y_t \) directly from \( y_0 \), \(a_\psi(x)\) and \( \epsilon \).

3. Rewrite \( q(y_{t-1}|y_t, y_0, x) \) using Eq.~\eqref{eq:bridge-posterior-noise-form}:

   \begin{equation}
   \tilde{\mu}_t(y_t, y_0, x)  = \frac{1}{\sqrt{1 - \beta_t}} \left( y_t - \frac{\beta_t}{\sqrt{1 - \bar{\alpha}_t}}\,\epsilon \right) + C(t)\,a_\psi(x).
   \end{equation}

4. Reparameterize \( \mu_\theta(y_t, t, x) \):
   Assume \( p_\theta(y_{t-1}|y_t, x) \) predicts \( \tilde{\mu}_t \):
   \begin{equation}
   \begin{aligned}
   \mu_\theta(y_t, t, x) &= \frac{1}{\sqrt{1 - \beta_t}}\left( y_t - \frac{\beta_t}{\sqrt{1 - \bar{\alpha}_t}}\epsilon_\theta(y_t, t, x) \right) \\
   &\quad + C(t)\,a_\psi(x)
   \end{aligned}
   \end{equation}
   Here, \( \epsilon_\theta(y_t, t, x) \) is a neural network predicting the noise \( \epsilon \).

5. Final Noise-Prediction Loss:
   The KL terms simplify to a weighted \( L_2 \) loss on the noise:
   \begin{equation}
   \mathbb{E}_{t, \epsilon} \left[ \frac{\beta_t^2}{2 \sigma_t^2 \alpha_t (1 - \bar{\alpha}_t)} \| \epsilon - \epsilon_\theta(y_t, t, x) \|^2 \right]
   \end{equation}
   Dropping the weighting (as in DDPM) gives the simplified loss:
   \begin{equation}
   \mathcal{L} = \mathbb{E}_{t, y_0, x, \epsilon} \left[ \| \epsilon - \epsilon_\theta(y_t, t, x) \|^2 \right],
   \end{equation}
   where:
   
   • \( t \sim \text{Uniform}(1, T) \),

   • \( y_t = \sqrt{\bar{\alpha}_t}\,y_0 + \bar{\gamma}_t\,a_\psi(x) + \sqrt{1 - \bar{\alpha}_t}\,\epsilon \).

This proves Eq.~\eqref{eq18}, and the denoising network \(\epsilon_\theta(y_t, t, x)\) is self-consistent with respect to the condition on \(x\).  

\subsection{Proof of Theorem~\ref{thm:wasserstein}}
\label{app:wasserstein}

We use the chain rule for KL divergence on Markov chains. For $t\geq 1$, the bridge transition is
\[
\begin{aligned}
&q(y_t\mid y_{t-1},x)\\
&\quad= \mathcal{N}\!\left(y_t;\,\sqrt{1-\beta_t}\,y_{t-1}+\gamma_t\,a_\psi(x),\,\beta_t I\right),
\end{aligned}
\]
which differs from $q(y_t\mid y_{t-1},x')$ only in its mean by $\gamma_t(a_\psi(x)-a_\psi(x'))$. Two Gaussians with shared covariance $\beta_t I$ and means $\mu_1,\mu_2$ have KL divergence $\frac{1}{2\beta_t}\|\mu_1-\mu_2\|_2^2$, so
\[
\begin{aligned}
&\mathrm{KL}\bigl(q(y_t\mid y_{t-1},x)\,\|\,q(y_t\mid y_{t-1},x')\bigr)\\
&\quad=\frac{\gamma_t^2}{2\beta_t}\bigl\|a_\psi(x)-a_\psi(x')\bigr\|_2^2,
\end{aligned}
\]
which is independent of $y_{t-1}$. Applying the Markov chain-rule for KL,
\[
\begin{aligned}
&\mathrm{KL}\!\left(\mathbb{P}^{\mathrm{NBP}}_{1:T\mid y_0,x}\,\big\|\,\mathbb{P}^{\mathrm{NBP}}_{1:T\mid y_0,x'}\right)\\
&\quad=\sum_{t=1}^T\mathbb{E}_{y_{t-1}\mid y_0,x}\!\left[\mathrm{KL}\bigl(q(\cdot\mid y_{t-1},x)\|q(\cdot\mid y_{t-1},x')\bigr)\right]\\
&\quad=\frac{1}{2}\!\left(\sum_{t=1}^T\frac{\gamma_t^2}{\beta_t}\right)\!\bigl\|a_\psi(x)-a_\psi(x')\bigr\|_2^2.
\end{aligned}
\]
For the marginal $W_2$ corollary, the forward marginal $q_t^{\mathrm{NBP}}(\cdot\mid y_0,x)=\mathcal{N}(\sqrt{\bar{\alpha}_t}y_0+\bar{\gamma}_t a_\psi(x),(1-\bar{\alpha}_t)I)$ shares its covariance across inputs, so $W_2$ between two such Gaussians equals the Euclidean distance between their means: $\bar{\gamma}_t\,\|a_\psi(x)-a_\psi(x')\|_2$. For NDP the forward kernel is independent of $x$, hence both the path KL and the marginal $W_2$ vanish. \hfill$\square$

\subsection{Proof of Theorem~\ref{thm:mutual-info}}
\label{app:mutual-info}

Write the NBP forward marginal in log-density form (additive constants in $y_t$ omitted):
\[
\begin{aligned}
&\log q_t^{\mathrm{NBP}}(y_t\mid y_0,x)\\
&\quad=-\frac{1}{2(1-\bar{\alpha}_t)}\bigl\|y_t-\sqrt{\bar{\alpha}_t}\,y_0-\bar{\gamma}_t\,a_\psi(x)\bigr\|_2^2+\mathrm{const}.
\end{aligned}
\]
Differentiating with respect to $x$ via the chain rule (with $J_\psi(x)=\partial a_\psi/\partial x\in\mathbb{R}^{D_y\times D_x}$),
\[
\begin{aligned}
&\nabla_x\log q_t^{\mathrm{NBP}}\\
&\quad=\frac{\bar{\gamma}_t}{1-\bar{\alpha}_t}\,J_\psi(x)^\top\!\bigl(y_t-\sqrt{\bar{\alpha}_t}\,y_0-\bar{\gamma}_t\,a_\psi(x)\bigr),
\end{aligned}
\]
proving Eq.~\eqref{eq:bridge-score}. Setting $\eta_t:=y_t-\sqrt{\bar{\alpha}_t}y_0-\bar{\gamma}_t a_\psi(x)\sim\mathcal{N}(0,(1-\bar{\alpha}_t)I)$ and using $\mathbb{E}[\eta_t\eta_t^\top]=(1-\bar{\alpha}_t)I$,
\[
\begin{aligned}
\mathcal{I}_t^{\mathrm{NBP}}(x)
&=\frac{\bar{\gamma}_t^2}{(1-\bar{\alpha}_t)^2}\,J_\psi(x)^\top\bigl((1-\bar{\alpha}_t)I\bigr)J_\psi(x)\\
&=\frac{\bar{\gamma}_t^2}{1-\bar{\alpha}_t}\,J_\psi(x)^\top J_\psi(x).
\end{aligned}
\]
For the path-level Fisher matrix in Eq.~\eqref{eq:fisher-path}, observe that the bridge is Markov in $t$: by the additivity of Fisher information across conditionally independent observations (here, conditional on $y_{t-1}$), and using the per-step transition $q(y_t\mid y_{t-1},x)$ analyzed in Sec.~\ref{app:wasserstein},
\[
\mathcal{I}_{1:T}^{\mathrm{NBP}}(x)
=\sum_{t=1}^T\mathcal{I}^{\mathrm{step}}_t(x)
=\left(\sum_{t=1}^T\frac{\gamma_t^2}{\beta_t}\right)J_\psi(x)^\top J_\psi(x).
\]
For NDP, the forward marginal does not contain $x$, so $\nabla_x\log q_t^{\mathrm{NDP}}\equiv 0$ and $\mathcal{I}_t^{\mathrm{NDP}}\equiv 0$. The Gaussian-prior corollary follows from the closed-form mutual information of a linear Gaussian channel: with $x\sim\mathcal{N}(0,\Sigma_x)$, $a_\psi=\mathrm{id}$ and $y_t=\sqrt{\bar{\alpha}_t}y_0+\bar{\gamma}_t x+\eta_t$ conditional on $y_0$, the mutual information equals $\frac{1}{2}\log\det(I+\bar{\gamma}_t^2(1-\bar{\alpha}_t)^{-1}\Sigma_x)$. \hfill$\square$

\subsection{Proof of Corollary~\ref{cor:cramer-rao}}
\label{app:cramer-rao}

Let $\hat{x}(y_{1:T})$ be any unbiased estimator of $x$ from the forward trajectory with $y_0$ fixed. The standard multi-parameter Cram\'er--Rao bound~\cite{hodges1951some,nielsen2013cramer} yields
\[
\mathrm{Cov}(\hat{x}(y_{1:T}))\succeq\bigl(\mathcal{I}_{1:T}^{\mathrm{NBP}}(x)\bigr)^{-1}.
\]
Substituting the path Fisher matrix from Theorem~\ref{thm:mutual-info},
\[
\mathrm{Cov}(\hat{x}(y_{1:T}))
\succeq\left(\sum_{t=1}^T\frac{\gamma_t^2}{\beta_t}\right)^{-1}\bigl(J_\psi(x)^\top J_\psi(x)\bigr)^{-1},
\]
which proves Eq.~\eqref{eq:cramer-rao} under the full-column-rank assumption on $J_\psi(x)$. For NDP the Fisher matrix is the zero matrix and the Cram\'er--Rao bound is degenerate; equivalently, since $y_{1:T}\perp\!\!\!\perp x\mid y_0$, no measurable estimator of $x$ from $y_{1:T}$ alone can be unbiased unless it is constant. \hfill$\square$

\subsection{Proof of Proposition~\ref{prop:direct-gradient}}
\label{app:direct-gradient}

The NBP noised state with anchor map is
\[
y_t(x)=\sqrt{\bar{\alpha}_t}\,y_0+\bar{\gamma}_t\,a_\psi(x)+\sqrt{1-\bar{\alpha}_t}\,\epsilon,
\qquad
\frac{\partial y_t}{\partial x}=\bar{\gamma}_t\,J_\psi(x).
\]
Applying the chain rule to $\ell_t(\theta,x)=\|\epsilon_\theta(y_t(x),x,t)-\epsilon\|_2^2$,
\[
\frac{d\ell_t}{dx}
=\frac{\partial \ell_t}{\partial x}
+\Bigl(\frac{\partial y_t}{\partial x}\Bigr)^\top\frac{\partial \ell_t}{\partial y_t}
=\frac{\partial \ell_t}{\partial x}
+\bar{\gamma}_t\,J_\psi(x)^\top\frac{\partial \ell_t}{\partial y_t},
\]
proving the chain-rule decomposition in Eq.~\eqref{eq:two-paths}; the explicit form $\partial \ell_t/\partial y_t = 2(\partial \epsilon_\theta/\partial y_t)^\top(\epsilon_\theta-\epsilon)$ follows from differentiating $\ell_t=\|\epsilon_\theta-\epsilon\|^2$ in $y_t$.

For the upper bound in Eq.~\eqref{eq:bridge-grad-magnitude}, write $r:=\epsilon-\epsilon_\theta(y_t,x,t)$ and $H_t:=\partial \epsilon_\theta(y_t,x,t)/\partial y_t$. Then
\[
\bar{\gamma}_t\,J_\psi(x)^\top\frac{\partial \ell_t}{\partial y_t}
= -2\bar{\gamma}_t\,J_\psi(x)^\top H_t^\top r,
\]
so by sub-multiplicativity of the operator norm,
\[
\Bigl\|\bar{\gamma}_t\,J_\psi(x)^\top\frac{\partial \ell_t}{\partial y_t}\Bigr\|^2
\;\leq\; 4\,\bar{\gamma}_t^2\,\|J_\psi(x)\|_{\mathrm{op}}^2\,\|H_t\|_{\mathrm{op}}^2\,\|r\|^2.
\]
Taking expectations gives the upper bound stated in the proposition. \emph{We do not claim equality}: the realized gradient magnitude depends jointly on the denoiser Jacobian $H_t$ and the residual $r$, neither of which the bridge construction prescribes; in particular, no general assumption on the spectrum, isotropy, or independence of $(H_t, r)$ is made. The Fisher-rate connection enters only through the prefactor $\bar{\gamma}_t^2$ and through normalizing $\mathbb{E}\|r\|^2$ by $(1-\bar{\alpha}_t)$, which is the expected residual scale at training optimum (under the standard DDPM noise covariance). For NDP, $\bar{\gamma}_t=0$ identically and the bridge term vanishes pointwise. \hfill$\square$

\paragraph{Schedule remark.} Under the SNR schedule $\gamma_t=\mathrm{SNR}_T/\mathrm{SNR}_t$, the cumulative coefficient is $\bar{\gamma}_t=\sum_{s=1}^t \gamma_s\sqrt{\bar{\alpha}_t/\bar{\alpha}_s}$. Even though $\gamma_s$ itself is monotone in $s$ for typical noise schedules, the rate $\bar{\gamma}_t^2/(1-\bar{\alpha}_t)$ is a ratio of two cumulative quantities and does not admit a closed-form monotonicity proof. We therefore verify numerically (Sec.~\ref{ablation}, Figure~\ref{fig:ablation_schedules}) that, for the cosine $\beta$-schedule used throughout this paper ($\beta_{\mathrm{start}}=3{\times}10^{-4}$, $\beta_{\mathrm{end}}=0.5$, $T=500$), the rate is monotonically increasing in $t$ and concentrated near the terminal phase. This is an empirical statement about the specific schedule we use, not a general analytical claim.

\section{Background: Neural Processes}
\label{np}
Neural Processes (NPs)~\cite{garnelo2018neural, garnelo2018conditional, kim2019attentive, louizos2019functional} combine the expressiveness of neural networks with the probabilistic reasoning of Gaussian Processes (GPs)~\cite{rasmussen2003gaussian}. While GPs offer principled uncertainty quantification, they suffer from poor scalability~\cite{snelson2005sparse, titsias2009variational} and limited kernel flexibility~\cite{wilson2016deep, liu2021deep}. In contrast, Neural Networks (NNs)~\cite{schmidhuber2015deep, nielsen2015neural} are highly flexible and scalable but lack inherent mechanisms for uncertainty modeling~\cite{blundell2015weight, pearce2020uncertainty, gawlikowski2023survey}. NPs address these limitations by modeling distributions over functions using a neural network-based framework. They approximate a stochastic process \( F: X \to Y \) through finite-dimensional marginals, parameterized by a latent variable \( z \) to capture global uncertainty. Given context observations \( (x_\mathcal{C}, y_\mathcal{C}) \) and target inputs \( x_\mathcal{T} \), NPs generate predictive distributions over \( y_\mathcal{T} \) via a conditional latent model.

\begin{equation}
p(y_\mathcal{T}, z|x_\mathcal{T}, x_\mathcal{C}, y_\mathcal{C}) = p(z|x_\mathcal{C}, y_\mathcal{C}) \prod_{i=1}^{|\mathcal{T}|} p(y_{\mathcal{T},i}|x_{\mathcal{T},i}, z)
\end{equation}

To ensure computational efficiency and order-invariance, NPs introduce three main components:
1. Encoder \( h \): Maps each context input-output pair \( (x_i, y_i) \) to a representation space, producing representations \( r_i = h(x_i, y_i) \).

2. Aggregator \( a \): Combines the encoded inputs into a single, permutation-invariant global representation \( r \). This is typically done by averaging the representations:$
r = a(r_i) = \frac{1}{|\mathcal{C}|} \sum_{i \in \mathcal{C}} r_i$. This global representation \( r \) parameterizes the latent distribution \( z \sim \mathcal{N}(\mu(r), I\sigma(r)) \).

3. Decoder \( g \): Predicts the target outputs \( y_\mathcal{T} = g(x_\mathcal{T}, z) \), conditioned on the latent variable \( z \) and the target inputs \( x_\mathcal{T} \).

Here, \( z \) encodes the uncertainty about the global structure of the underlying function. Training NPs uses amortized variational inference, optimizing an evidence lower bound (ELBO) on the conditional log-likelihood:
\begin{equation}
\begin{aligned}
&\log p(y_\mathcal{T}|x_\mathcal{C}, y_\mathcal{C}, x_\mathcal{T}) \\
&\geq \mathbb{E}_{q(z|x_\mathcal{C}, y_\mathcal{C},x_\mathcal{T}, y_\mathcal{T})} \Big[ \sum_{i \in \mathcal{T}} \log p(y_{\mathcal{T},i}|z, x_{\mathcal{T},i}) \\
&\qquad\qquad + \log \frac{p(z|x_\mathcal{C}, y_\mathcal{C})}{q(z|x_\mathcal{C}, y_\mathcal{C},x_\mathcal{T}, y_\mathcal{T}))} \Big]
\end{aligned}
\end{equation}
where \( q(z|x_\mathcal{C}, y_\mathcal{C},x_\mathcal{T}, y_\mathcal{T}) \) is the variational posterior distribution parameterized by a neural network, and \( p(z|x_\mathcal{C}, y_\mathcal{C}) \) is the conditional prior.

\section{NDP Review}

Formally, given a function $f: \mathbb{R}^D \to \mathbb{R}$, an NDP learns a generative distribution over observed data pairs $(x,y)$, where inputs $x \in \mathbb{R}^{N \times D}$ and outputs $y = f(x) \in \mathbb{R}^N$. Unlike standard NPs, NDPs \cite{dutordoir2023neural} do not explicitly require a partitioning into context and target sets during training; all points are jointly modeled. In supervised learning setting, the NDP modeling framework consists of two stochastic processes:

\paragraph{Forward Diffusion Process.}
Starting from observed clean data $y_0$, the forward diffusion process gradually injects Gaussian noise into the outputs over $T$ timesteps according to a predefined variance schedule $\{\beta_t\}$:
\begin{equation}
\begin{aligned}
q(y_{1:T} \mid y_0) &= \prod_{t=1}^T q(y_t \mid y_{t-1}),\\
q(y_t \mid y_{t-1}) &= \mathcal{N}(y_t; \sqrt{1 - \beta_t} \, y_{t-1}, \beta_t I).
\end{aligned}
\end{equation}
After $T$ diffusion steps, the distribution of the outputs converges towards standard Gaussian noise, i.e., $y_T \sim \mathcal{N}(0, I)$. 

\paragraph{Reverse  Process.}  
Neural Diffusion Processes (NDPs) learn a conditional reverse process that denoises observations from Gaussian noise $ y_T $ to outputs $ y_0 $, guided by an input $ x $:  
\begin{equation}  
p_\theta(y_{0:T} \mid x) = p(y_T) \prod_{t=1}^T p_\theta(y_{t-1} \mid y_t, x),  
\end{equation}  
with Gaussian transitions parameterized by a noise prediction model $ \epsilon_\theta $:  
\begin{equation}  
p_\theta(y_{t-1} \mid y_t, x) = \mathcal{N}\left(y_{t-1}; \mu_\theta(y_t, t, x), \tilde{\beta}_t I\right).  
\end{equation}

where $
 \tilde{\beta}_t = \frac{1 - \bar{\alpha}_{t-1}}{1 - \bar{\alpha}_t} \beta_t$,
$\mu_\theta(y_t, t, x)$ is reparameterized as \begin{equation}
\begin{aligned}
\mu_\theta(y_t, t, x) &= \frac{1}{\sqrt{1 - \beta_t}} \left( y_t - \frac{\beta_t}{\sqrt{1 - \bar{\alpha}_t}} \epsilon_\theta(y_t, t, x) \right),\\
\bar{\alpha}_t &= \prod_{s=1}^t (1 - \beta_s)
\end{aligned}
\end{equation}

\paragraph{Training Objective.}
NDPs employ a denoising score matching objective ~\cite{hyvarinen2005estimation,song2021maximum,huang2021variational}, training the noise model $\epsilon_\theta$ by minimizing the discrepancy between predicted noise and actual noise $\epsilon \sim \mathcal{N}(0, I)$:
\begin{equation}
\begin{aligned}
\mathcal{L}_\theta &= \mathbb{E}_{t, x, y_0, \epsilon} \left[ \left\| \epsilon - \epsilon_\theta(y_t, t, x) \right\|_2^2 \right],\\
\text{with} \quad y_t &= \sqrt{\bar{\alpha}_t} y_0 + \sqrt{1 - \bar{\alpha}_t} \epsilon.
\end{aligned}
\end{equation}

\paragraph{Conditional Sampling Procedure.}
At test time,  NDPs draw samples from the conditional distribution $p(y_{\mathcal{T},0} \mid x_{\mathcal{T}}, D)$, where $D = (x_\mathcal{C} \in \mathbb{R}^{M \times D}, y_{\mathcal{C},0} \in \mathbb{R}^M)$ is the context observations.

The conditional sampling proceeds as follows. First, sample the initial target noise: $
y_{\mathcal{T},T} \sim \mathcal{N}(0, I).
$ For each diffusion timestep $t = T, \ldots, 1$, proceed with:

\begin{itemize}
    \item Sample the noisy version of the context points using the forward diffusion process:
    \begin{equation}
    y_{\mathcal{C},t} \sim \mathcal{N}\left( \sqrt{\bar{\alpha}_t} y_{\mathcal{C},0}, (1 - \bar{\alpha}_t) I \right),
    \end{equation}
    
    \item Form the combined dataset at time $t$ by collecting the union of noisy targets and noisy contexts:
    \begin{equation}
    y_t = \{ y_{\mathcal{T},t}, y_{\mathcal{C},t} \}, \quad x = \{ x_{\mathcal{T}}, x_\mathcal{C}\}.
    \end{equation}
    
    \item Perform the reverse denoising step by sampling from the learned backward kernel:
    \begin{equation}
    \begin{aligned}
    y_{t-1} &\sim \mathcal{N} \left( \frac{1}{\sqrt{\alpha_t}} \left( y_t - \frac{\beta_t}{\sqrt{1 - \bar{\alpha}_t}} \epsilon_\theta(y_t, t,x) \right), \sigma_t^2  I \right),\\
    &\quad \text{where} \quad \alpha_t = 1 - \beta_t.
    \end{aligned}
    \end{equation}
   
\end{itemize}
\begin{table}[h]
\centering
\caption{Comparison of generation properties.}
\begin{tabular}{lcc}
\textbf{Method} & Endpoint Match & Path Consistency \\
\hline
NDP (Baseline) & Implicit & Weak \\
NBP (Ours) & \checkmark & \checkmark \\
\end{tabular}
\end{table}
\section{Noise Model Architecture }
\label{nma}

To ensure that our model remains consistent with the structural properties of stochastic processes and to guarantee fair experimental comparisons, we adopt the same noise model architecture as NDPs, namely the Bi-Dimensional Attention Block~\cite{dutordoir2023neural}. 

\begin{figure*}[htbp]
    \centering
    \includegraphics[width=\textwidth]{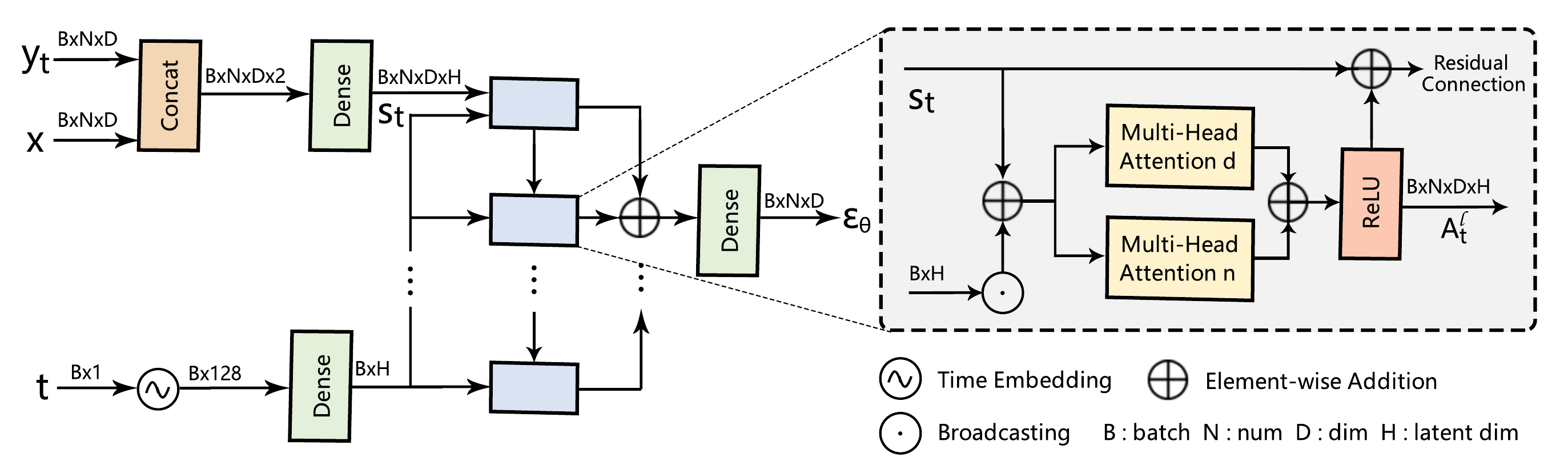}
    \caption{Noise model architecture employed at each step of Neural Bridge Processes. The greyed-out box highlights the bi-dimensional attention block inherited from NDP.}
    \label{fig:example}
\end{figure*}

As described in Figure \ref{fig:example}, this architecture is designed to encode two key symmetries:

\begin{itemize}
    \item \textbf{Exchangeability over data points:} the model should be equivariant to permutations of the dataset ordering. That is, shuffling the order of inputs in the context or target set should not affect the output distribution.
    \item \textbf{Invariance over input dimensions:} the prediction should be unaffected by reordering of input features (e.g., swapping the order of columns in a tabular dataset).
\end{itemize}

To accommodate both properties, the Bi-Dimensional Attention Block operates over a tensor $s_t \in \mathbb{R}^{N \times D \times H}$ representing the latent representation of paired inputs $(x, y_t)$ and timestep $t$. Each block consists of two multi-head self-attention (MHSA) mechanisms:
\begin{itemize}
    \item $\text{MHSA}_N$: acts across the \emph{dataset axis} $N$, propagating information across data points;
    \item $\text{MHSA}_D$: acts across the \emph{input dimension axis} $D$, capturing interactions between features.
\end{itemize}

The output of each block at layer $\ell$ is computed as:
\[
\begin{aligned}
&A_t^{(\ell)}(s_t^{(\ell-1)})\\
&\quad = A_t^{(\ell-1)} + \sigma\left( \text{MHSA}_N(s_t^{(\ell-1)}) + \text{MHSA}_D(s_t^{(\ell-1)}) \right),
\end{aligned}
\]
where $\sigma$ denotes a ReLU activation, and $A_t^{(0)} = 0$, $s_t^{(0)} = s_t$ is the output of the preprocessing stage.

Each Bi-Dimensional Attention Block maintains equivariance under permutations of data and feature dimensions:
\begin{proposition}[Equivariance~{[Prop. 4.1]\cite{dutordoir2023neural}}]
Let $\pi_N$ and $\pi_D$ be permutations over dataset and feature axes respectively. Then,
\[
\pi_D \circ \pi_N \circ A_t(s) = A_t(\pi_D \circ \pi_N \circ s), \quad \forall s \in \mathbb{R}^{N \times D \times H}.
\]
\end{proposition}

The final noise model $\epsilon_\theta$ is obtained by summing outputs across all Bi-Attention layers, followed by an aggregation over the input dimension axis to remove dependence on feature order. This leads to:

\begin{proposition}[ Equivariance~{[Prop. 4.2]\cite{dutordoir2023neural}}]
Let $\pi_N, \pi_D$ be permutations as above. Then,
\[
\pi_N \circ \epsilon_\theta(x_t, y_t, t) = \epsilon_\theta(\pi_N \circ \pi_D \circ x_t, \pi_N \circ y_t, t).
\]
\end{proposition}

By encoding these structural properties, NDP and NBP generate, at each step, a set of exchangeable random variables— a structural requirement commonly used in NDP models to construct consistent finite-dimensional conditional distributions. However, as with NDP, our method does not attempt to satisfy the strict marginal consistency conditions required by the Kolmogorov Extension Theorem, even though the structure allows the model to empirically approximate consistency during meta-learning.

\section{More Details for Experiments}

\label{details}
\subsection{Complete Experimental Protocol}
\label{sec:full-protocol}

Tables~\ref{tab:protocol-train}--\ref{tab:protocol-eval} give the training, architecture and evaluation settings in full, so that every number in the paper can be reproduced without reference to prior work. Each baseline uses its own reference architecture under the shared optimization protocol of Table~\ref{tab:protocol-train}; NBP inherits NDP's backbone and budget. We did not run separate hyperparameter sweeps per baseline, and we state this as a restriction on the scope of the empirical claims rather than presenting the configurations as tuned optima.

\begin{table}[t]
\centering
\caption{Training protocol, identical for every model on the Gaussian-process tasks.}
\label{tab:protocol-train}
\setlength{\tabcolsep}{4pt}
\resizebox{\columnwidth}{!}{%
\begin{tabular}{@{}ll@{}}
\toprule
Item & Value \\
\midrule
Optimizer & Adam \\
Learning rate & $5\times10^{-4}$, cosine-annealed to $0$ \\
Training steps & $100\,000$ \\
Batch size (tasks per step) & 16 \\
Latent samples per step & 4 (latent-variable models) \\
Context / target sampling & $n_{\mathrm{ctx}}\!\sim\!\mathcal{U}(3,N\!-\!3)$, $n_{\mathrm{tar}}\!\sim\!\mathcal{U}(3,N\!-\!n_{\mathrm{ctx}})$, $N=50$ \\
Input range & $x\sim\mathcal{U}(-2,2)^{D}$ \\
Data & GP prior, RBF and Matérn-5/2; lengthscale $\sim\mathcal{U}(0.1,0.6)\cdot\sqrt{D}$, \\
 & output scale $\sim\mathcal{U}(0.1,1.0)$, observation noise $2\times10^{-2}$ \\
Seeds & 5 per configuration (3 for CylinderFlow), paired across models \\
Hardware & NVIDIA RTX A6000 \\
\bottomrule
\end{tabular}}
\end{table}

\begin{table}[t]
\centering
\caption{Architectures. All baselines use their reference configurations.}
\label{tab:protocol-arch}
\setlength{\tabcolsep}{4pt}
\resizebox{\columnwidth}{!}{%
\begin{tabular}{@{}ll@{}}
\toprule
Model & Architecture \\
\midrule
NBP / NDP (ours) & identical transformer denoiser: $d_{\mathrm{model}}$ 64, 6 layers, 4 heads, \\
 & FFN 128, embedding depth 4, positional-encoding dim 20, \\
 & $T=100$ diffusion steps, dropout 0; the bridge flag is the \\
 & \emph{only} difference \\
CNP & mean-pooling encoder (pre-depth 4, post-depth 2), decoder \\
 & depth 3, hidden 128 \\
CANP & cross-attention encoder (v-depth 4, qk-depth 2) $+$ the CNP \\
 & encoder, decoder depth 3, hidden 128 \\
NP & CNP encoder $+$ latent path, $\dim_{\mathrm{lat}}$ 128, $\sigma$-bound 0.05 \\
ANP & CANP encoder $+$ latent path, $\dim_{\mathrm{lat}}$ 128, $\sigma$-bound 0.05 \\
TNP-A & transformer, $d_{\mathrm{model}}$ 64, 6 layers, 4 heads, FFN 128, \\
 & permutation-augmented \\
FlowNP & same transformer backbone as ours, 100 integration steps \\
ConvCNP / ConvGNP & official \texttt{neuralprocesses} package; UNet ($32$ channels $\times5$), \\
 & points-per-unit $64/24/12$ for $D=1/2/3$; likelihood \texttt{het} / \\
 & \texttt{lowrank} respectively \\
GP reference & exact posterior of the generating kernel \\
\bottomrule
\end{tabular}}
\end{table}

\begin{table}[t]
\centering
\caption{Evaluation protocol.}
\label{tab:protocol-eval}
\setlength{\tabcolsep}{4pt}
\resizebox{\columnwidth}{!}{%
\begin{tabular}{@{}ll@{}}
\toprule
Item & Value \\
\midrule
Evaluation batches $\times$ batch size & $2\times100$ tasks \\
Conditional samples per task & 128 (diffusion / flow models) \\
Likelihood estimator & Gaussian fitted to the samples, full empirical \\
(diffusion / flow) & covariance $+\,10^{-2}$ jitter, reported \emph{per target point} \\
Likelihood estimator & the model's own analytic predictive density, \\
(CNP/NP family, TNP, Conv) & per target point \\
NDP conditional sampler & official RePaint, \texttt{num\_inner\_steps} $=5$ \\
CylinderFlow & 20 evaluation batches, MSE in raw velocity units \\
Bayesian optimization & 20 independent tasks, 200-candidate pool, \\
 & 15 iterations, Thompson sampling, simple regret \\
\bottomrule
\end{tabular}}
\end{table}

\subsection{Baselines in Autoregressive Mode and Under a Common Estimator}
\label{sec:ar-baselines}

Conditional members of the CNP family factorize over target points and therefore cannot represent correlations within a single forward pass. Following~\cite{bruinsma2023autoregressive}, we also evaluate them autoregressively, using the chain rule with teacher forcing, which yields a valid joint density. Table~\ref{tab:ar-mode} shows that AR evaluation improves every conditional baseline we tested: all $12/12$ runs improve, with an overall mean gain of $+0.197\pm0.060$~nat at $D=1$. AR-CNP is therefore included as a baseline rather than treated as a weaker variant of CNP.

\begin{table}[t]
\centering
\caption{Autoregressive evaluation improves every conditional CNP baseline. Target log-likelihood per target point ($\uparrow$) at $D=1$, mean $\pm$ standard deviation over 3 seeds.}
\label{tab:ar-mode}
\setlength{\tabcolsep}{4pt}
\begin{tabular}{@{}lccc@{}}
\toprule
Model & Standard & AR mode & Gain \\
\midrule
CNP (RBF) & $0.304\pm0.092$ & $0.496\pm0.038$ & $+0.192\pm0.059$ \\
CANP (RBF) & $0.711\pm0.077$ & $0.939\pm0.017$ & $+0.228\pm0.063$ \\
CNP (Matérn) & $0.092\pm0.048$ & $0.254\pm0.012$ & $+0.162\pm0.042$ \\
CANP (Matérn) & $0.447\pm0.096$ & $0.653\pm0.048$ & $+0.206\pm0.055$ \\
\bottomrule
\end{tabular}
\end{table}

Table~\ref{tab:tnp} reports the transformer Neural Process baseline. TNP-A uses its own analytic predictive density, so its values are not on the same scale as the Gaussian-fit estimates of the diffusion models and we do not draw a ranking across the two protocols.

\begin{table}[t]
\centering
\caption{TNP-A target log-likelihood ($\uparrow$), per-seed values under its own analytic predictive density.}
\label{tab:tnp}
\begin{tabular}{@{}ll@{}}
\toprule
Setting & Seeds \\
\midrule
$D=1$ (RBF) & 1.68 / 1.87 / 1.35 \\
$D=1$ (Matérn) & 1.19 / 1.48 / 1.03 \\
$D=2$ (RBF) & 0.376 / 0.206 \\
$D=3$ (RBF) & $-0.443$ / $-0.384$ \\
\bottomrule
\end{tabular}
\end{table}

\subsection{Details of EEG Dataset Regression Task}
\begin{figure*}[htbp]
    \centering
    \includegraphics[width=0.95\textwidth]{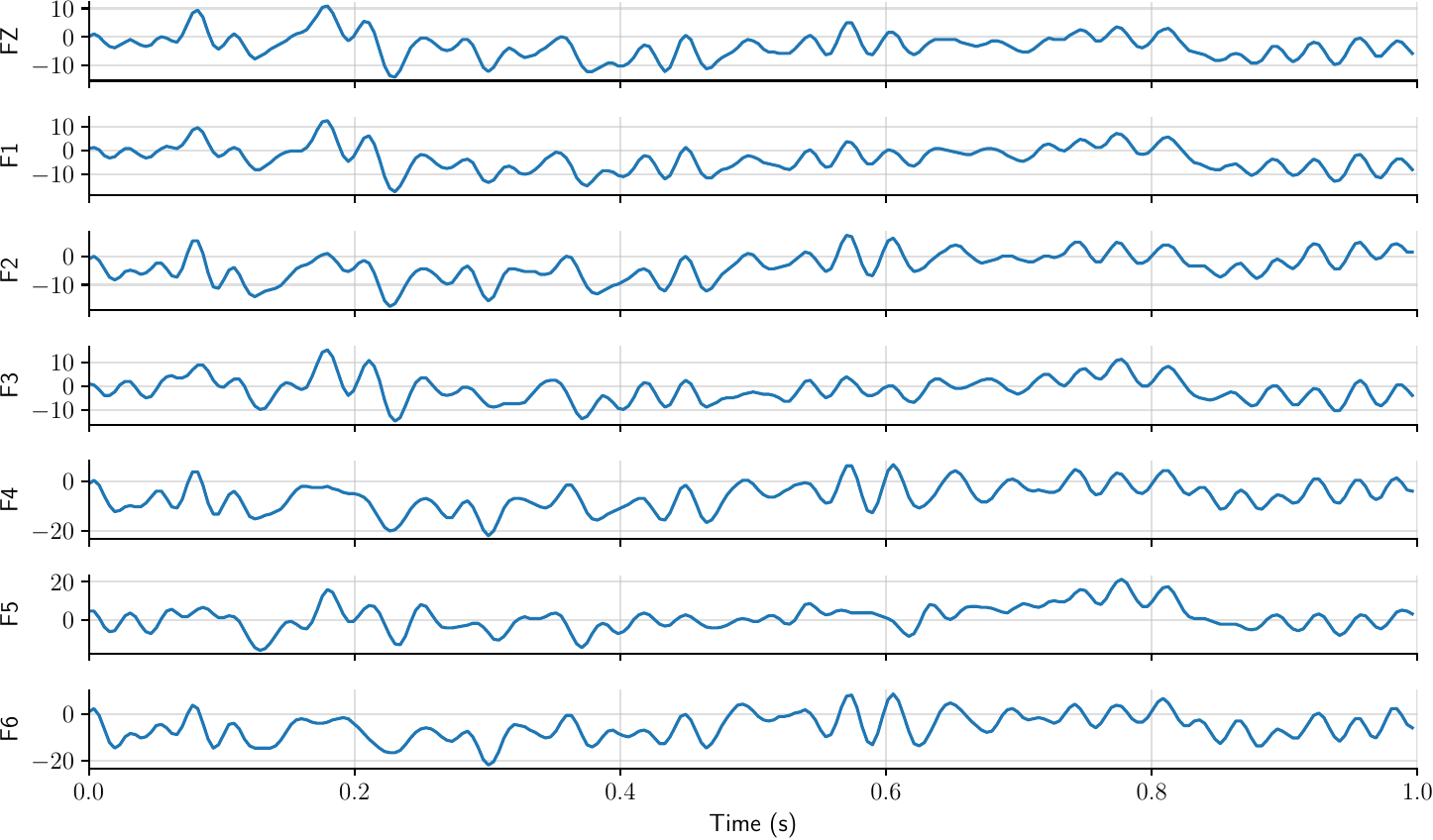}
    \caption{the signals recorded from these seven channels for a single trial of one subject.}
    \label{fig:2}
\end{figure*}

The EEG dataset used in this experiment  consists of recordings from 122 subjects, including both alcoholic and control groups. Each subject underwent either single or double stimulus conditions, during which neural responses were recorded using 64 scalp electrodes. Each trial lasted for 1 second with a sampling rate of 256 Hz, and up to 120 trials were recorded per subject.

For our study, we focus on signals from 7 frontal electrodes: FZ, F1, F2, F3, F4, F5, and F6. This selection yields a total of 7,632 multivariate time series, each comprising 256 time steps across 7 channels. These signals exhibit strong temporal dynamics and inter-channel correlations, making the dataset well-suited for evaluating the generalization and modeling capabilities of multi-output meta-learning models. The data is publicly available from the UCI Machine Learning Repository, with collection details described in~\cite{zhang1995event}. Figure \ref{fig:2} illustrates the signals recorded from these seven channels for a single trial of one subject.

Subjects were split into training, validation, and test sets on a per-individual basis. The validation and test sets each contain 10 subjects, with the remainder assigned to the training set. All trials from each subject form a single meta-task, enabling task-level generalization evaluation.

Within each trial, we randomly select 3 out of the 7 channels and mask partial segments of these channels to simulate missing data. This setup supports three complementary prediction regimes. Interpolation evaluates whether the model can recover locally missing values within the observed time range. Reconstruction asks the model to infer masked regions of a target channel using the remaining channels as context, thereby testing cross-channel dependency modeling. Forecasting evaluates extrapolation to future signal trajectories based on currently observed measurements. Each input is represented as an index vector $\mathbf{x}_e = (i_t, i_c)$, where $i_t$ denotes the time step and $i_c$ the channel index. The corresponding output is the voltage signal $\mathbf{y}_e$.

All models were trained for 1,000 iterations on the training set. Evaluation metrics include Mean Squared Error (MSE) and Negative Log-Likelihood (NLL). The Neural Bridge Process (NBP) employs a 5-layer Bi-Dimensional Attention Block with hidden dimension 64 and 8 attention heads. We train and evaluate each model over multiple random seeds and report the mean performance, using the same seed set for all compared methods.

For models incorporating diffusion-based generation (such as NDP and NBP), we adopt a cosine noise schedule with the following parameters: $\beta_{\text{start}} = 0.0003$, $\beta_{\text{end}} = 0.5$, and 500 diffusion timesteps. These settings are applied consistently across the forward and reverse processes in all diffusion-based models. Additional training hyperparameters are as follows:
\begin{itemize} 
\item For NBP and NDP base, the learning rate was set to $2 \cdot 10^{-5}$;

\item Other models used default learning rates as recommended in prior literature;

\item All models operated on input sequences of 256 time steps.
\end{itemize} 
The evaluation results, summarized in Table~\ref{tab:eeg_results} of the main text, show that NBP obtains lower NLL and MSE than the compared baselines across the three EEG prediction tasks.

\subsection{Details of Image Regression Task}

\begin{figure*}[ht]
    \centering
    \resizebox{0.8\textwidth}{!}{
        \begin{minipage}{\textwidth}
            \begin{subfigure}{0.32\textwidth}
                \centering
                \includegraphics[width=\linewidth]{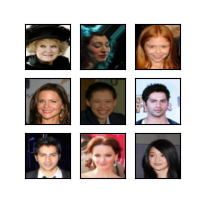}
                \caption{Ground Truth}
                \label{fig:ground_truth}
            \end{subfigure}
            \hfill
            \begin{subfigure}{0.32\textwidth}
                \centering
                \includegraphics[width=\linewidth]{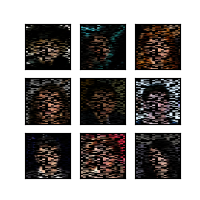}
                \caption{Corrupted Context}
                \label{fig:corrupted_context}
            \end{subfigure}
            \hfill
            \begin{subfigure}{0.32\textwidth}
                \centering
                \includegraphics[width=\linewidth]{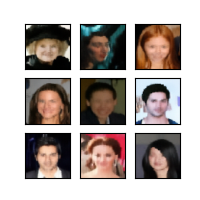}
                \caption{Model Prediction}
                \label{fig:model_prediction}
            \end{subfigure}
        \end{minipage}
    }
    \caption{Qualitative results on the CelebA 64$\times$64 image regression task.}
    \label{fig:celeba_results}
\end{figure*}

We provide detailed information on the image regression task using Neural Bridge Processes (NBPs). The task involves predicting pixel intensities based on their spatial coordinates, which are normalized to the range $[-2, 2]$. We use the CelebA dataset at resolutions of $32 \times 32$ and $64 \times 64$.

Our experimental protocol—including the denoising network architecture, training schedule, optimizer configuration, and seed set—closely follows the setup used for Neural Diffusion Processes (NDPs), ensuring a fair and consistent comparison. The core architecture of the NBP model consists of 7 layers, each with hidden dimension 64 and 8 attention heads. Sparse attention is not used in these experiments.

\paragraph{Training Configuration.} The model is trained using a batch size of 32. The optimizer follows the same learning-rate schedule as the NDP baseline:
\begin{itemize}
\item Initial learning rate: $2.0 \times 10^{-5}$
\item Peak learning rate: $1.0 \times 10^{-3}$
\item End learning rate: $1.0 \times 10^{-5}$
\item Warmup and cosine decay are applied following the NDP implementation.
\item EMA decay rate: 0.995
\end{itemize}

\paragraph{Diffusion Settings.} We employ a cosine beta schedule with the following parameters for the forward and reverse processes:
\begin{itemize}
\item $\beta_{\text{start}} = 0.0003$
\item $\beta_{\text{end}} = 0.5$
\item Number of timesteps: 500
\end{itemize}

\paragraph{Evaluation Protocol.} Each prediction is averaged over 9 conditional samples during testing. The evaluation batch size is set to 9, with 128 samples drawn per image for final averaging. All pixel values are normalized to the $[0, 1]$ range. Results are averaged over multiple random seeds using the same seed set across compared methods.

\paragraph{Loss Function.} We adopt the $\ell_1$ loss for training the denoising objective.

\paragraph{Results and Analysis.} Tables~\ref{t1} and~\ref{t2} report the quantitative performance under various levels of context sparsity. NBP obtains a lower mean MSE than NDP at every reported setting. For example, at a context ratio of 0.2 on CelebA $32 \times 32$, NBP achieves an MSE of 0.77 compared with 0.88 for NDP, and on CelebA $64 \times 64$ an MSE of 0.86 compared with 1.05 under the same sparse-context condition. As noted in the main text, the per-ratio standard deviations are of the same order as these gaps, so the evidence here is the consistency of the ordering across nine context ratios and two resolutions rather than a per-ratio significant difference; the likelihood comparison in Table~\ref{tab:celeba_nll} is the sharper of the two.

These results are consistent with the proposed design, in which the forward diffusion kernel is explicitly conditioned on the input coordinates through the bridge term. The component ablation in Sec.~\ref{sec:celeba_component_ablation} further separates this effect from endpoint initialization, learned coordinate-anchor capacity, and denoiser capacity.

\section{Quantifying Path Supervision: Sensitivity Analysis}
\label{sec:sensitivity}

A key advantage of NBP over NDP lies in its \textit{input-anchored forward kernel} (Eq.~8 and~10 in the main text), which explicitly injects input $x$ into the entire diffusion trajectory rather than relying solely on the denoiser to implicitly learn conditional guidance. To quantify this advantage, we conduct a sensitivity analysis measuring how strongly the denoiser $\epsilon_\theta(y_t, x, t)$ depends on the input $x$ at different timesteps.

\paragraph{Measurement: Jacobian Sensitivity.}
For a given sample, we compute the sensitivity of the denoiser output with respect to input $x$ at each timestep $t$:
\begin{equation}
S(t) = \mathbb{E}\left[\left\|\frac{\partial \epsilon_\theta(y_t, x, t)}{\partial x}\right\|_F\right]
\end{equation}
where $\|\cdot\|_F$ denotes the Frobenius norm. In PyTorch, we enable \texttt{requires\_grad=True} for $\epsilon_\theta$, compute the norm of gradients with respect to $x$, and then normalize both the output gradient and input gradient to obtain relative sensitivity. This metric directly measures the \textit{effective coupling} between input and output through the learned denoiser.

\paragraph{Hypothesis.}
We expect the following behavior:
\begin{itemize}
    \item \textbf{NDP}: $S(t)$ should increase gradually from early to late timesteps, approaching 0 in early phases (because the forward process does not contain $x$, relying solely on implicit "soft conditioning" through the denoiser).
    \item \textbf{NBP}: $S(t)$ should remain substantial across \textit{all} timesteps, especially in \textbf{early phases}, reflecting the explicit injection of $x$ into the forward kernel via $\gamma_t x$.
\end{itemize}

\paragraph{Experimental Setup.}
We evaluate on CelebA $32 \times 32$ with context ratio 0.3. For each method (NDP and NBP), we:
\begin{enumerate}
    \item Sample 100 test images and randomly mask 70\% of pixels.
    \item For each timestep $t \in \{1, 50, 100, 150, \ldots, 500\}$, forward diffuse the image to obtain $y_t$.
    \item Compute $S(t)$ by averaging the Jacobian norm $\|\partial \epsilon_\theta / \partial x\|_F$ over all samples.
\end{enumerate}

\paragraph{Results.}
Figure~\ref{fig:sensitivity_curve} in the main text presents the sensitivity curves for NDP and NBP on CelebA \(32\times32\) with context ratio 0.3. In this diagnostic, NDP has lower measured sensitivity in early timesteps and increases later, while NBP shows larger early-timestep sensitivity. Across the sampled timesteps with \(t\leq150\), the measured \(S(t)\) for NBP is roughly 3--5\(\times\) larger than that of NDP. This is consistent with the proposed mechanism that the bridge term gives the denoiser an additional input-dependent training state, but we interpret it as supporting evidence rather than a standalone guarantee.

\paragraph{Interpretation.}
The elevated sensitivity $S(t)$ in NBP's early timesteps stems from the explicit $\gamma_t a_\psi(x)$ term in the forward kernel. Because $y_t$ itself depends on $x$, the loss responds to the input along two routes rather than one:
\begin{enumerate}
    \item \textbf{Direct path}: $x \to y_t \to \epsilon_\theta$ (via forward kernel)
    \item \textbf{Indirect path}: $x \to \epsilon_\theta$ (via denoiser conditioning)
\end{enumerate}
In NDP the first route is absent, so the full burden falls on the denoiser-side conditioning mechanism. These are routes of \emph{sensitivity} to the input, not separate supervision signals: training updates $(\theta,\psi)$, not $x$. The sensitivity measurement is consistent with the presence of the additional route in NBP, especially when the signal-to-noise ratio is high (early \(t\)).

\paragraph{Implications.}
These diagnostics suggest several possible explanations for the empirical improvements:
\begin{itemize}
    \item \textbf{Optimization}: Larger early-phase gradients may make input-output dependencies easier to learn.
    \item \textbf{Trajectory conditioning}: Input-dependent training states reduce the reliance on the denoiser-side conditioning pathway alone.
    \item \textbf{Prediction quality}: The bridge path provides an additional mechanism for aligning generated outputs with the observed input structure.
\end{itemize}

In summary, the sensitivity analysis provides diagnostic evidence that NBP changes where input information enters the diffusion model. We do not treat this metric as a direct performance measure; rather, it complements the quantitative results and ablations by illustrating the proposed path-level conditioning mechanism.

%%%%%%%%%%%%%%%%%%%%%%%%%%%%%%%%%%%%%%%%%%%%%%%%%%%%%%%%%%%%

% NeurIPS checklist removed (not applicable to IEEE journals):
% \input{checklist.tex}

\end{document}